\documentclass[a4paper,conference]{IEEEtran}

\usepackage{amsmath}
\usepackage{graphicx}
\usepackage{threeparttable}
\usepackage{dcolumn}
\usepackage{booktabs}
\usepackage{amsmath,graphicx}
\usepackage{url}
\usepackage{cite}
\usepackage{psfrag}
\usepackage{amsfonts,amssymb}
\usepackage{amsbsy}
\usepackage{graphicx}
\usepackage{epstopdf}
\usepackage{array}
\usepackage{multirow}
\usepackage{bm}
\usepackage{subfigure}
\usepackage{verbatim}
\usepackage{colortbl}
\usepackage{stfloats}
\usepackage{algorithm}
\usepackage{algorithmic}
\usepackage{setspace}

\usepackage{bigstrut}
\usepackage{amsmath}
\usepackage{mathtools}

\begin{document}

\title{An Extension of LIME with Improvement of Interpretability and Fidelity}
%\title{Toward Local Explanation: Feature Correlation Sampling and Nonlinear Approximation}

% \author{\IEEEauthorblockN{Sheng Shi\IEEEauthorrefmark{1}, Yangzhou Du\IEEEauthorrefmark{1},
% Xinfeng Zhang\IEEEauthorrefmark{2} and
% Wei Fan\IEEEauthorrefmark{1}}
% \IEEEauthorblockA{\IEEEauthorrefmark{1}AI Laboratory, Lenovo Research, Beijing 100094, China\\ Email: \{shisheng2, duyz1 and fanwei2\}@lenovo.com}
% \IEEEauthorblockA{\IEEEauthorrefmark{2}University of Chinese Academy of Sciences, Beijing 100049, China\\ Email: xfzhang@ucas.ac.cn}}

\author{\IEEEauthorblockN{Sheng Shi,
Yangzhou Du and
Wei Fan}
\IEEEauthorblockA{AI Laboratory, Lenovo Research, Beijing 100094, China\\ Email: \{shisheng2, duyz1 and fanwei2\}@lenovo.com}}

\maketitle

\begin{abstract}
While deep learning makes significant achievements in Artificial Intelligence (AI), the lack of transparency has limited its broad application in various vertical domains. Explainability is not only a gateway between AI and real world, but also a powerful feature to detect flaw of the models and bias of the data. Local Interpretable Model-agnostic Explanation (LIME) is a widely-accepted technique that explains the prediction of any classifier faithfully by learning an interpretable model locally around the predicted instance. As an extension of LIME, this paper proposes an high-interpretability and high-fidelity local explanation method, known as Local Explanation using feature Dependency Sampling and Nonlinear Approximation (LEDSNA). Given an instance being explained, LEDSNA enhances interpretability by feature sampling with intrinsic dependency. Besides, LEDSNA improves the local explanation fidelity by approximating nonlinear boundary of local decision. We evaluate our method with classification tasks in both image domain and text domain. Experiments show that LEDSNA's explanation of the back-box model achieves much better performance than original LIME in terms of interpretability and fidelity.

\end{abstract}

% \fi
\IEEEpeerreviewmaketitle

\section{Introduction}
In recent years, people have witnessed the fast development of Artificial Intelligence (AI) \cite{AI01,AI02,AI03}. Compared to traditional machine learning methods, deep learning  has achieved superior performance in many challenging tasks. There has been an increasing interest in leveraging deep learning methods to aid decision makers in critical domains such as healthcare and criminal justice. However, because of the nested complicated structure, deep learning models remain mostly black boxes, which are extremely weak in explaining the reasoning process and prediction results. This makes it challenging for decision makers to understand and trust their functionality. Therefore, the explainability and transparency of deep learning models are essential to ensure their broad applications in various vertical domains.

Recently, the development of techniques on explainability and transparency of deep learning models has recently received much attention in the research community \cite{EX01,EX02,EX03}. Among them, the post-hoc techniques for explaining black-box models in a human-understandable manner have received much attention \cite{POST01,POST02,POST03}, which generate perturbed samples of a given instance in the feature space and observe the effect of these perturbed samples on the output of the black-box classifier. Due to the generality, these techniques have been used to explain neural networks and complex ensemble models in various domains ranging from medicine, law and finance \cite{App01} \cite{App02}. The most representative system in this category is LIME\cite{POST01}. As LIME assumes the local area of the classification boundary near the input instance is linear, it uses a linear regression model which is self-explanatory to locally represent the decision and pinpoint important features based on the regression coefficients. It is found relevant works \cite{Other01,Other02,Other03} proposed to use other models such as decision tree to approximate the target detection boundaries.

There are two drawbacks in current existing local explanations such as LIME. Perturbed samples are generated from a uniform distribution, ignoring the intrinsic correlation between features. This may lead to lose much useful information to learn the local explanation models. Proper sampling operation is especially essential in natural language processing and image recognition. Moreover, most existing methods assume the decision boundary is local linearity, which may produce serious errors as in most complex networks, the local decision boundary is non-linear.

In this paper, we design and develop a novel, high-interpretability and high-fidelity local explanation method to address the above challenges. First, we design a unique local sampling process which incorporate the feature clustering method to handle the feature dependency problems. Then, we adopt Support Vector Regression (SVR) with a kernel function to approximate locally nonlinear boundary. In this way, by simultaneously preserving feature dependency and local non-linearity, our method produces high-interpretability and high-fidelity explanation. For convenience, we refer to our method as LEDSNA ``Local Explanation using feature Dependency Sampling and Nonlinear Approximation''.

\section{Method}
In this section, we first introduce the two core characteristics of the local explanation method: interpretability and fidelity. Then we introduce the feature sampling with intrinsic dependency and nonlinear boundary of local decision. Finally, we present the framework of LEDSNA algorithm.

% \subsection{Tradeoff between fidelity and understandability}
An explainable model with good interpretability should be faithful to the original model, understandable to the observer, and graspable in a short time so that the end-user can make wise decisions. Local explanation method learns a model from a set of data samples which is sampled around the instance being explained. The dissimilarity between the true label and predicted label is defined as the loss function $L(f(x),g(x))$ which is a measure of how unfaithful $g(x)$ is in approximating $f(x)$. In order to ensure both local fidelity and understandability, we add regularization term to loss function:
%L0-norm
\begin{equation}
J(\theta)=argmin{L(f(x), g_{\theta}(x)) + \lambda\Omega(\theta)}.
%J(\theta)=argmin{L(f(x), g_{\theta}(x)) + {\lVert\theta\rVert}_0}.
\end{equation}
The regularisation term is a measure of complexity of the explainable model $g(x)$. The smaller the regularisation term is, the better the sparsity of model $g(x)$, which leads to better understandability. This is the general framework of LIME \cite{POST01}.
%We choose L0-norm as regularisation term, L0-norm is the number of non-zero parameters which are learned in approximation process. It is a measure of interpretability, that is to say the smaller the number of non-zero parameters is, the better the understandability.

\subsection{Feature Sampling with Intrinsic Dependency}
In current existing local explanations, the original sampling procedure is made on each feature independently, ignoring the intrinsic correlation between features. Proper sampling operation is essential as the independent sampling process may lead to lose much useful information to learn the local explanation models. In some cases, when most uniformly generated samples are unrealistic about the actual distribution, false information contributors lead to poorly fitting of the local explanation model. In this section, we design an unique local sampling process which incorporate the feature clustering method to activate a subset of features for better local exploration.

%According to the feature dependence of image and text, we incorporate different clustering methods in the sampling process.
%In this section, in order to learn the local behavior of classifier $f$, we generate a group of perturbed samples of a given instance, $x$, by activating a subset of features in $x$.
%the features especially for image exhibit a strong correlation in the spacial neighborhood.

%we design unique local sampling process to solve feature dependency problem.

\subsubsection{Feature Dependency Sampling for Image}

\begin{figure}
\begin{minipage}{0.24\linewidth}
  %\centering
  \centerline{\includegraphics[width=1.0\textwidth]{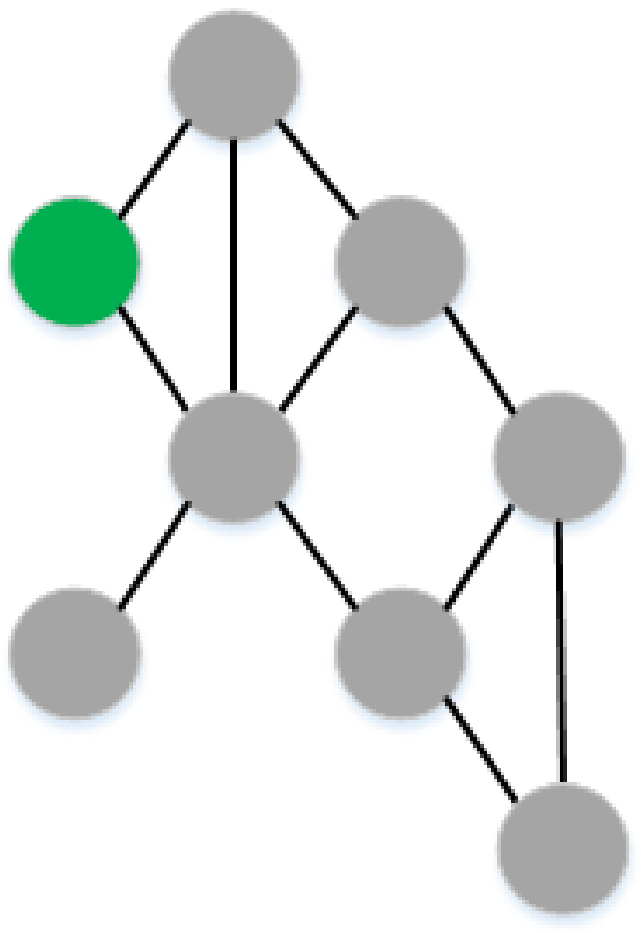}}
  %\centerline{\scriptsize{(a)}}
  %\centerline{}
\end{minipage}
\hfill
\begin{minipage}{0.24\linewidth}
  %\centering
  \centerline{\includegraphics[width=1.0\textwidth]{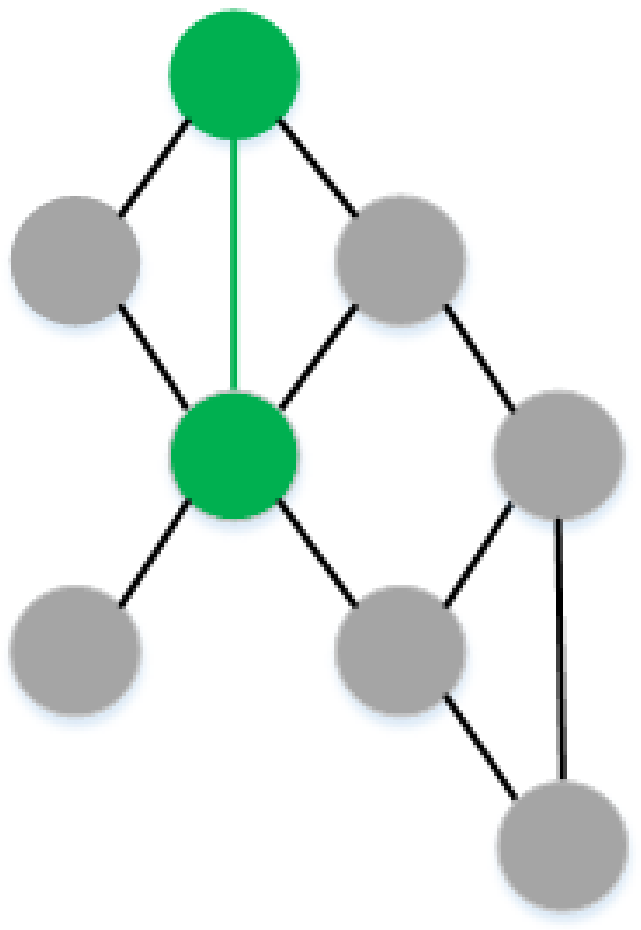}}
  %\centerline{\scriptsize{(b)}}
  %\centerline{}
\end{minipage}
\hfill
\begin{minipage}{0.24\linewidth}
  %\centering
  \centerline{\includegraphics[width=1.0\textwidth]{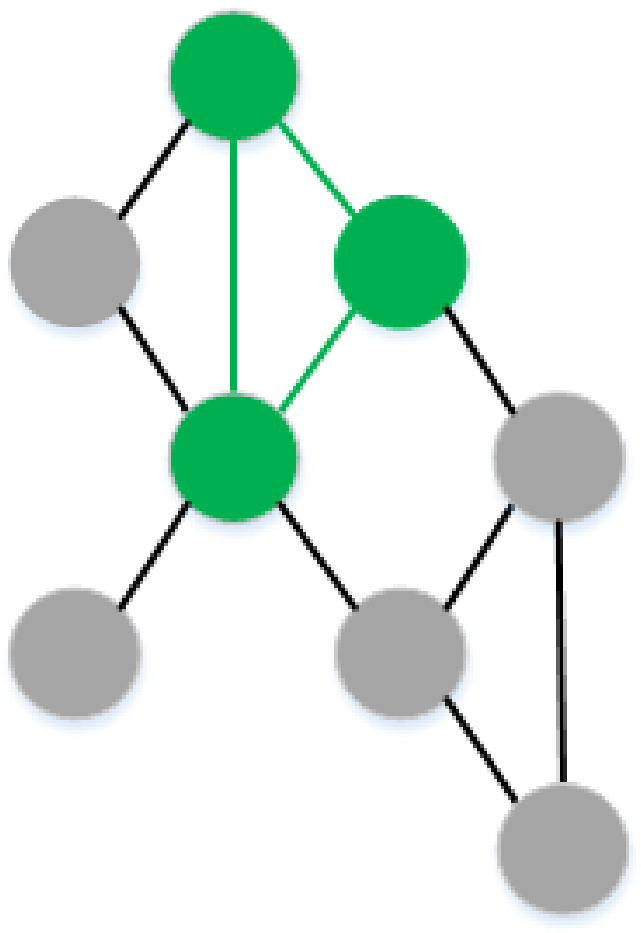}}
  %\centerline{\scriptsize{(c)}}
  %\centerline{}
\end{minipage}
\hfill
\begin{minipage}{0.24\linewidth}
  %\centering
  \centerline{\includegraphics[width=1.0\textwidth]{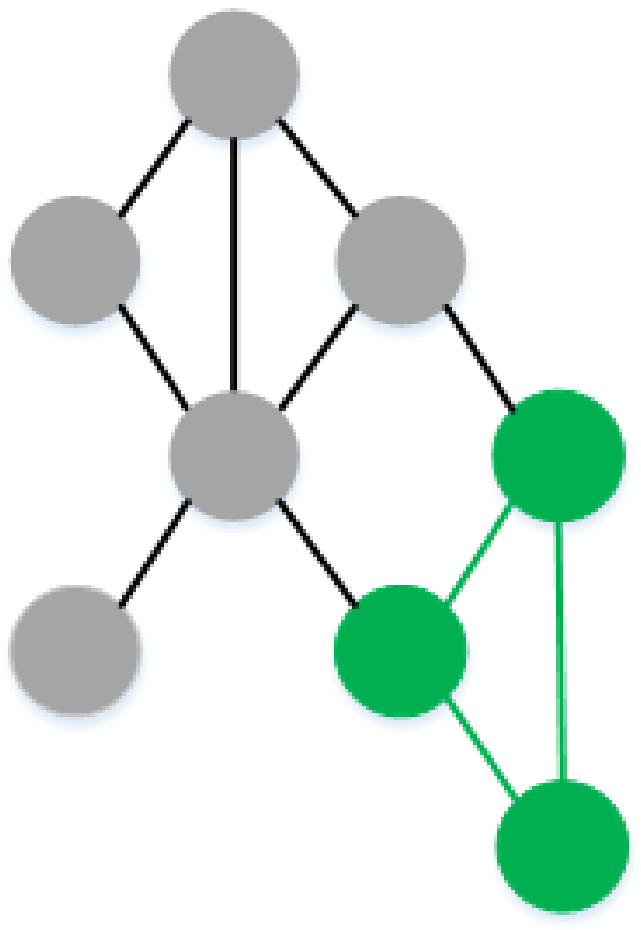}}
  %\centerline{\scriptsize{(c)}}
  %\centerline{}
\end{minipage}
\caption{\small{Some samples in the clique $C$, where every two vertices are adjacent (marked green)}}
\label{fig:2}
\end{figure}

Proper sampling operation is especially essential in natural image recognition because the visual features of natural objects exhibit a strong correlation in the spacial neighborhood. For image classification, we adopt a superpixel based interpretable representation. Each superpixel segment is the primary processing unit, which is a group of connected pixels with similar colors or gray levels. We denote $x\in{\mathbb{R}^d}$ be the original representation of an image, and binary vector $x'\in{\{{0,1}\}^{d'}}$ be its interpretable representation, which indicating the ¡®presence¡¯ or ¡®absence¡¯ of a superpixel segment. There $d$ is the number of pixels and $d'$ is the number of superpixel. For the images, especially natural images, superpixel segments often correspond to the coherent regions of visual objects, showing strong correlation in a spacial neighborhood. In order to learn the local behavior of image classifier $f$, we generate a group of perturbed samples of a given instance, $x$, by activating a subset of superpixels in $x$. Firstly, we convert the superpixel segments into an undirected graph. the superpixel segments are represented as vertices of a graph whose edges connect to only those adjacent segments. Considering a graph $G=(V, E)$, where $V$ and $E$ are the sets of vertices and undirected edges, with cardinalities $|V|=d'$ and $|E|$, a subset of $V$ can be represented by a binary vector $z'\in{\{{0,1}\}^{d'}}$, where $1$ indicates that vertice is in the subset. The perturbed sampling operation is formalized as finding the clique $C$ ($C\subseteq V$), where every two vertices are adjacent. We use the Depth-First Search (DFS) method to get the clique $C$. Some samples in the clique are shown in Fig.~\ref{fig:2}. Since there is a strong correlation between the adjacent superpixel image segments, the clique $C$ set construction can take into full account the various types of neighborhood correlation.
\begin{figure}
\begin{minipage}{0.32\linewidth}
  %\centering
  \centerline{\includegraphics[width=1.0\textwidth]{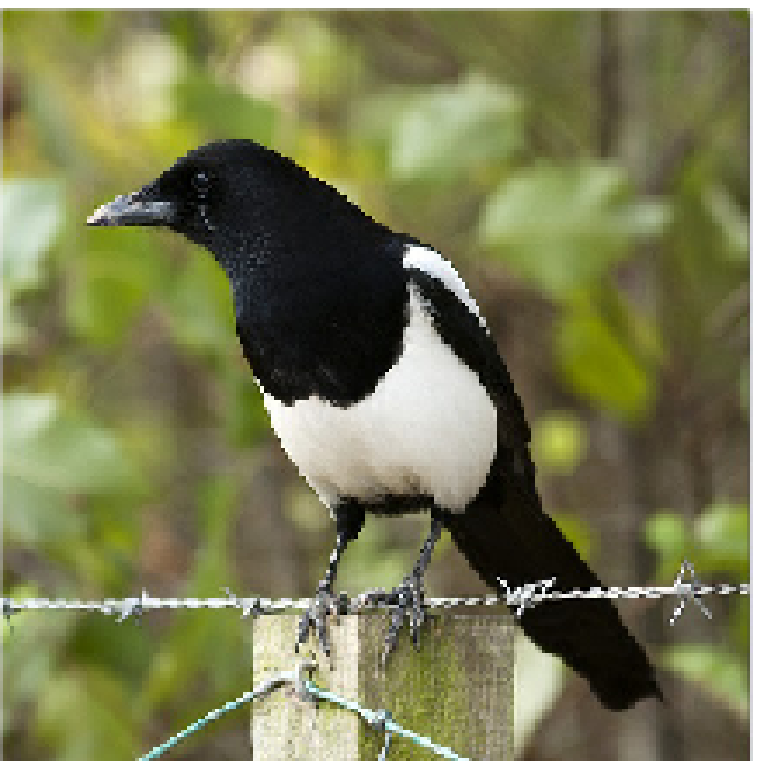}}
  \centerline{\scriptsize{(a)}}
  %\centerline{}
\end{minipage}
\hfill
\begin{minipage}{0.32\linewidth}
  %\centering
  \centerline{\includegraphics[width=1.0\textwidth]{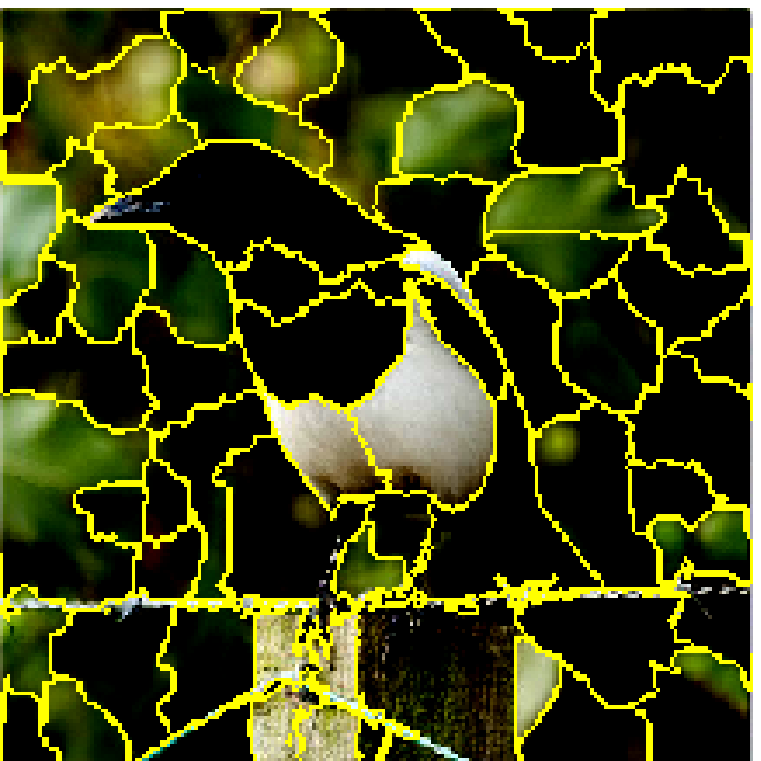}}
  \centerline{\scriptsize{(b)}}
  %\centerline{}
\end{minipage}
\hfill
\begin{minipage}{0.32\linewidth}
  %\centering
  \centerline{\includegraphics[width=1.0\textwidth]{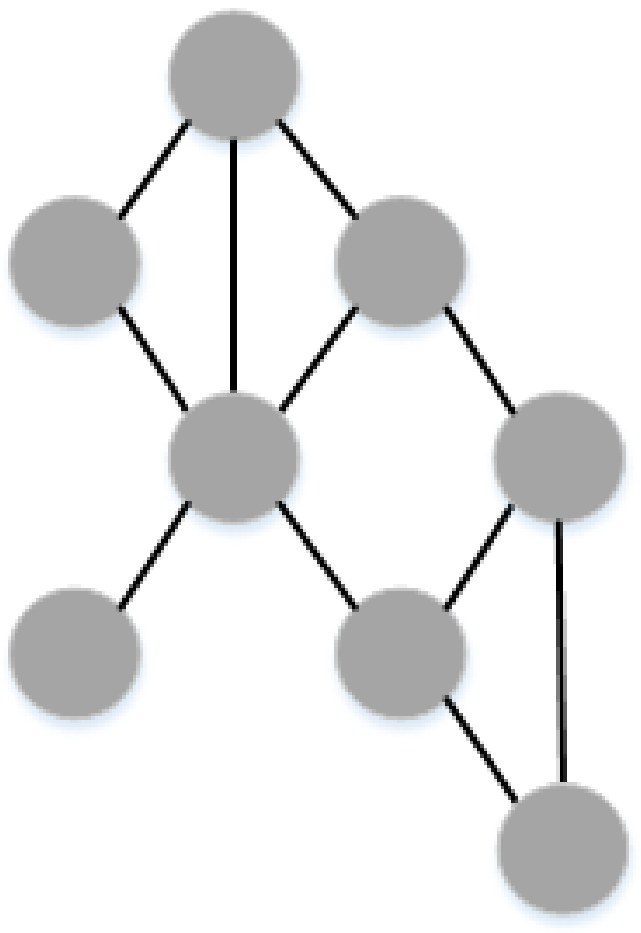}}
  \centerline{\scriptsize{(c)}}
  %\centerline{}
\end{minipage}
\caption{\small{(a) Pixel-based image; (b) Superpixel image; (c) Constructing a graph of all superpixel blocks}}
\label{fig:1}
\end{figure}

\subsubsection{Feature Dependency Sampling for Text}

It is also essential for natural language processing to have a proper sampling operation. For text classification, we let the interpretable representation be a bag of words. Similar to image, $x\in{\mathbb{R}^d}$ denotes the original representation of a text, and binary vector $x'\in{\{{0,1}\}^{d'}}$ denotes its interpretable representation. In order to learn the local behavior of text classifier, we generate a group of perturbed samples of a given instance by activating a subset of features. Fig.~\ref{fig:3} shows two natural language in Chinese and English, we can find there are strong semantic dependency between words especially in Chinese. If the activated features are get by using a sampling process where features are independent to each other, we may loss much useful information to learn the local explanation models. In sampling process, the semantic dependent words correspond to adjacent superpixels in the image. Semantic dependent words should be selected or unselected at the same time. There are many methods to analyze semantic dependency of natural language. There, we incorporate the Stanford CoreNLP \cite{CoreNLP} tools into sampling process to get the perturbed samples.
% \begin{figure}
% \centering
% \begin{minipage}{0.8\linewidth}
%   \centering
%   \centerline{\includegraphics[width=1.0\textwidth]{figures/chi-seg}}
%   %\centerline{\scriptsize{(a)}}
%   %\centerline{}
% \end{minipage}
% \caption{\small{Semantic dependency of text}}
% \label{fig:3}
% \end{figure}

% \begin{figure}
% \centering
% \begin{minipage}{0.9\linewidth}
%   \centering
%   \centerline{\includegraphics[width=1.0\textwidth]{figures/eng-seg}}
%   %\centerline{\scriptsize{(a)}}
%   %\centerline{}
% \end{minipage}
% \caption{\small{Semantic dependency of English natural language}}
% \label{fig:4}
% \end{figure}

\begin{figure}
\centering
\begin{minipage}{0.7\linewidth}
  \centering
  \centerline{\includegraphics[width=1.0\textwidth]{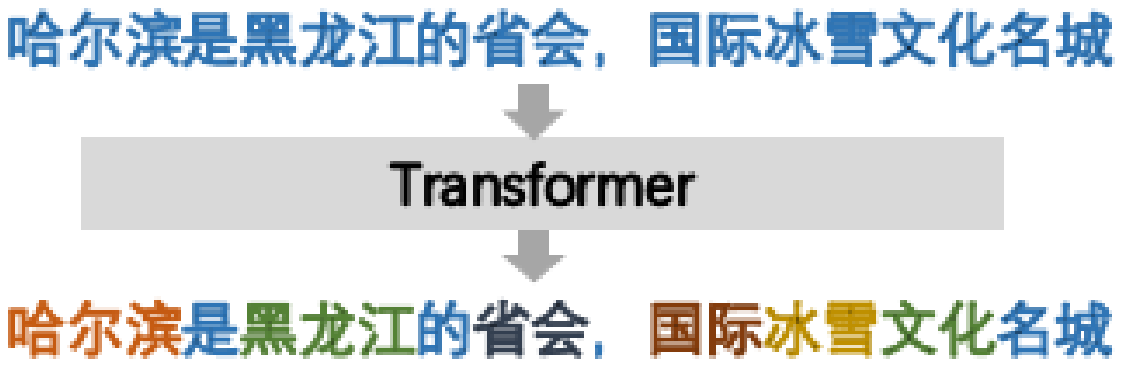}}
  \centerline{\scriptsize{(a)}}
  \centerline{}
\end{minipage}
\vfill
\begin{minipage}{0.9\linewidth}
  \centering
  \centerline{\includegraphics[width=1.0\textwidth]{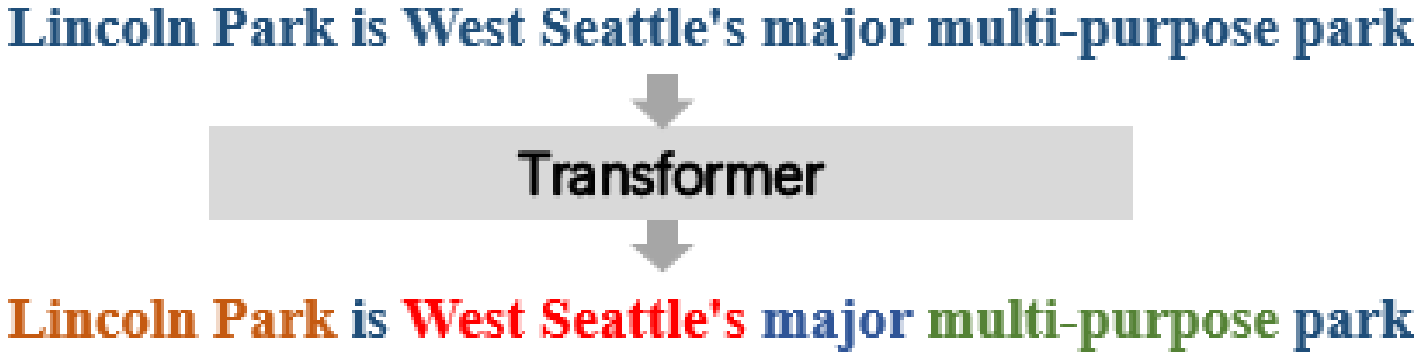}}
  \centerline{\scriptsize{(b)}}
  \centerline{}
\end{minipage}
\caption{\small{Semantic dependency of Chinese natural language and English natural language}}
\label{fig:3}
\end{figure}
%independent semantic

\subsection{Nonlinear Boundary of Local Decision }
Most existing local explanation methods assume the decision boundary is local linearity. Those explanation methods may produce serious errors as in most complex networks, the local decision boundary is non-linear. Experiments show a simple linear approximation will significantly degrade the explanation fidelity. In this section, we adopt Support Vector Regressor (SVR) with kernel function to approximate nonlinear boundary. In approximation processing, when data are not distributed linearly in the current feature space, we use kernel function to project data points into higher dimensional feature space and find the optimal hyperplane.

The perturbed samples of a given instance are impossible to be fitted by a linear model. Our way to tackle this problem is to apply a kernel function mapping to bring data to a higher dimensional feature space.
% The Gaussian kernel function computes the similarity between the data points in a much higher dimensional space:
% \begin{equation}
% k(x,x')=e^{-(x-x')^2/\sigma^2}.
% \end{equation}
The formula to transfrom the data is as follow:
\begin{equation}
g(\boldsymbol{x},\boldsymbol{w})=\sum_{i=1}^N{{w_i}k(x-x')}.
%g(x,\boldsymbol{w})=\sum_{i=1}^N{{w_i}e^{-(x-x')^2/\sigma^2}}=\boldsymbol{w}^\top\Phi(x).
\end{equation}
% Function for regression $g(x,\boldsymbol{w})$ is a non-linear function of $\boldsymbol{x}, but linear in $\boldsymbol{w}$
% \begin{figure}
% \centering
% \begin{minipage}{0.8\linewidth}
%   \centering
%   \centerline{\includegraphics[width=1.0\textwidth]{figures/guassian}}
%   %\centerline{\scriptsize{(a)}}
%   %\centerline{}
% \end{minipage}
% \caption{\small{Guassian kernel project data points into higher dimensional feature space and find the optimal hyperplane}}
% \label{fig:4}
% \end{figure}
After project data point into higher dimensional feature space. We search for a hyperplane by using hinge error measure. Specifically, we introduce slack variables for data points that violate $\varepsilon-$insensitive error:
\begin{equation}
\nonumber err(f(x_i),(g(x_i,\boldsymbol{w}))=
\end{equation}
\begin{equation}
\begin{cases}
0,& \|f(x_i)-g(x_i,\boldsymbol{w})\|\leqslant{\varepsilon}\\
\|f(x_i)-g(x_i,\boldsymbol{w})\|-{\varepsilon}, & \|f(x_i)-g(x_i,\boldsymbol{w})\|>{\varepsilon}
\end{cases}
\end{equation}

\begin{figure}
\centering
\begin{minipage}{0.8\linewidth}
  \centering
  \centerline{\includegraphics[width=0.8\textwidth]{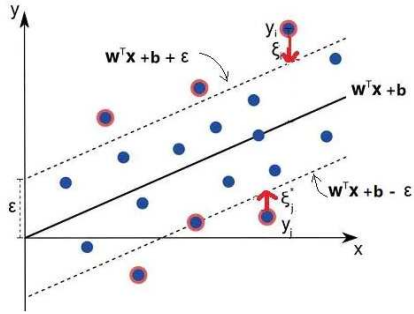}}
  %\centerline{\scriptsize{(a)}}
  %\centerline{}
\end{minipage}
\caption{\small{Two slack variables are required to measure the distance between points and tube}}
\label{fig:5}
\end{figure}
For each data point $x_i$, two slack variables, $\xi_i, \hat{\xi_i}$ are required to measure whether $g(x_i)$ is above or below the tube.
\begin{equation}
\begin{cases}
\xi_i=f(x_i)-(g(x_i,\boldsymbol{w})+\varepsilon),& if f(x_i)>g(x_i,\boldsymbol{w})+\varepsilon\\
\xi_i=0,& otherwise
\end{cases}
\end{equation}

\begin{equation}
\begin{cases}
\hat{\xi_i}=(g(x_i,\boldsymbol{w})-\varepsilon)-f(x_i),& if f(x_i)<g(x_i,\boldsymbol{w})-\varepsilon\\
\hat{\xi_i}=0,& otherwise
\end{cases}
\end{equation}
The learning is by the optimization:
\begin{eqnarray}
\nonumber &\mathop{\min}_{\boldsymbol{w},\xi_i,\hat{\xi_i}}{\sum_{i=1}^N(\xi_i+\hat{\xi_i})+\lambda{\lVert\boldsymbol{w}\rVert}^2}\\
\nonumber  s.t:\\ \nonumber & f(x_i)-g(x_i,\boldsymbol{w})\geqslant{\varepsilon+\xi_i}; \\ \nonumber &f(x_i)-g(x_i,\boldsymbol{w})\leqslant{\varepsilon+\hat{\xi_i}};\\
& \xi_i\geqslant0; \hat{\xi_i}\geqslant0, i=1,2,...,N.
\end{eqnarray}
This is the famous support vector regression method which can be solved by building Lagrangian functions.

Algorithm 1 shows a simplified workflow diagram of LEDSNA. Firstly, LEDSNA incorporates the feature clustering method into sampling process to activate a subset of features. Then, LEDSNA uses kernel function to project data points into higher dimensional feature space. Finally, LEDSNA use the support vector regression to search for a hyperplane and get the coefficient of important feature.
\begin{algorithm}[h]
\caption{\small{Local Explanation using feature Dependency Sampling and Nonlinear Approximation (LEDSNA)}}
\label{alg::conjugateGradient}
\begin{algorithmic}[1]
\footnotesize
\REQUIRE
Classifier $f$,
%Number of samples $N$;
Instance $x$,
%Length of explanation $K$
%Max depth of tree $d$;
%\ENSURE
%time and prediction error of TLIME;
\STATE get interpretable presentation of $x'$ (e.g. superpixel image for image and bag of word for text)
\STATE get $f(x')$ by classifier $f$
\STATE incorporate the feature clustering method into sampling process to activate a subset of features
\STATE initial $Z \leftarrow \{\}$
% \STATE construct the clique $C$ by DFS method
%\STATE get $z'$ which is  $z'\in C$;
\FOR {$z'\in C$}
    \STATE get $z$ by recovering $z'$
    \STATE $Z \leftarrow Z \cup (z'_i,f(z_i),\pi_x(z_i))$
\ENDFOR
%\STATE output $\omega$ $\leftarrow$ K-Lasso(Z, K);
\STATE use kernel function to project data points into higher dimensional feature space: $g(\boldsymbol{x},\boldsymbol{w})=\sum_{i=1}^N{{w_i}k(x-x')}.$;
\STATE use the support vector regression to search for a hyperplane
\RETURN  feature coefficient
\end{algorithmic}
\end{algorithm}
%\begin{equation}
%\hat{J}(\boldsymbol{w})=argmin{\sum_{i=1}^{N}{\{f(x_i)-g(x_i,\boldsymbol{w})\}^2} + \lambda{\lVert\boldsymbol{w}\rVert}^2}.
%%J(\theta)=argmin{L(f(x), g_{\theta}(x)) + {\lVert\theta\rVert}_0}.
%\end{equation}

\section{Experiments}
In this section, we first introduce the evaluation criterion of explanation methods. Then, we perform experiments in natural language processing in Chinese. Finally, we perform experiment to explain the Google's pre-trained Inception neural network \cite{Inception} on imagenet database. Experiment results show the flexibility of LEDSNA.

\subsection{Evaluation criterion}
A good explainable model requires same characteristics. One of the essential criterion is interpretability. The explanation must appear as a certain form understandable to the observer, i.e., providing visual explanations which lists most significant features contributed to the prediction.

Another essential criterion is local fidelity. The explanation must be faithful to the model in the vicinity of the instance being predicted. Local Approximation Error ($Err$) and R-squared ($R^2$) are two important measurements of the accuracy of our local approximation with respect to the original decision boundary. Local Approximation Error can reflect the prediction accuracy:
\begin{equation}
Err=|f(x_0)-g(x_0)|,
\end{equation}
where $f(x_0)$ is a single prediction obtained from a target deep learning classifier, $g(x_0)$ is the predicted value by explanation model. $R^2$ is the ``percent of variance explained'' by the explanation model. That is to say that $R^2$ is the fraction by which the variance of the errors is less than the variance of the dependent variable. $R^2$ is calculated by Total Sum of Squares ($SST$) and Error Sum of Squares ($SSE$):
\begin{eqnarray}
\nonumber & R^2=1-SSE/SST\\
\nonumber& SSE=\sum_{i=1}^n(f(x_i)-g(x_i))^2\\
& SST=\sum_{i=1}^n(f(x_i)-f_{mean})^2,
\end{eqnarray}
where $f(x_i)$ is the label of perturbed sample $x_i$, obtained from a target deep learning classifier. $g(x_i)$ is the predicted value and $f_{mean}$ is the mean value of $f(x_i)$. Moreover, $R^2$ can be expressed by Mean Square Error ($MSE$) and Variance ($Var$) which are familiar to us:
\begin{eqnarray}
% \nonumber & R^2=1-{{\frac{1}{n}}\sum_{i=1}^n(f(x_i)-g(x_i))^2}/{{\frac{1}{n}}\sum_{i=1}^n(f(x_i)-f_{mean})^2}\\
\nonumber & R^2=1-\frac{{\frac{1}{n}}\sum_{i=1}^n(f(x_i)-g(x_i))^2}{{\frac{1}{n}}\sum_{i=1}^n(f(x_i)-f_{mean})^2}\\
& =1-MSE/Var
%& =1-\frac{MSE}{Var}
\end{eqnarray}
$R^2$ is a relative measure which is conveniently scaled between 0 and 1. The best $R^2$ is $1.0$. The closer the score is to $1.0$, the better the performance of fidelity is to explainer.

\subsection{Experiment on Image Classifiers}
%\subsection{Google's Inception neural network on Image-net database}

\begin{figure}
\begin{minipage}{0.24\linewidth}
  %\centering
  \centerline{\includegraphics[width=1.0\textwidth]{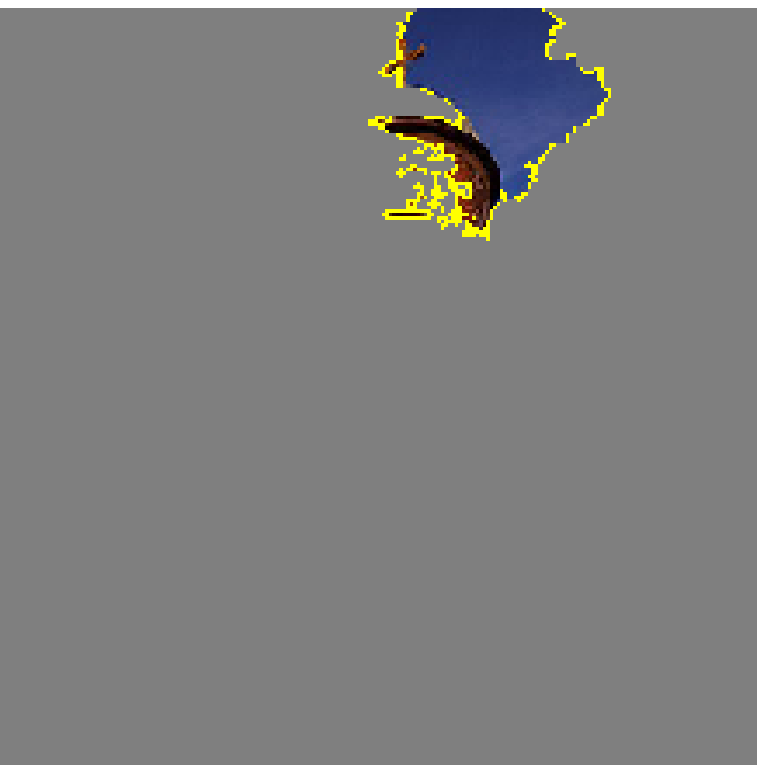}}
  \centerline{\scriptsize{(a) LIME (K=1)}}
  %\centerline{}
\end{minipage}
\hfill
\begin{minipage}{0.24\linewidth}
  %\centering
  \centerline{\includegraphics[width=1.0\textwidth]{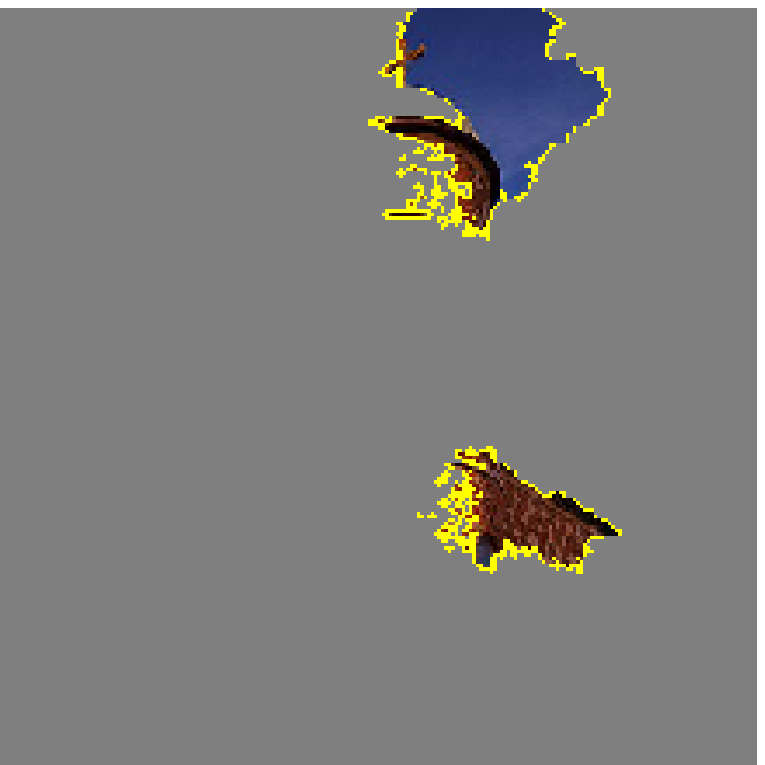}}
  \centerline{\scriptsize{(b) LIME (K=2)}}
  %\centerline{}
\end{minipage}
\hfill
\begin{minipage}{0.24\linewidth}
  %\centering
  \centerline{\includegraphics[width=1.0\textwidth]{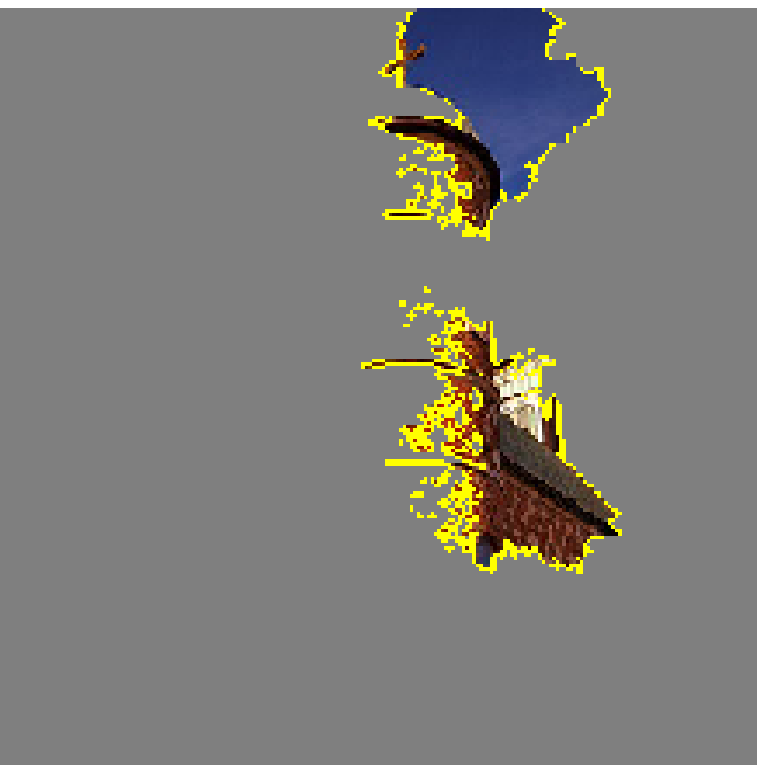}}
  \centerline{\scriptsize{(c) LIME (K=3)}}
  %\centerline{}
\end{minipage}
\hfill
\begin{minipage}{0.24\linewidth}
  %\centering
  \centerline{\includegraphics[width=1.0\textwidth]{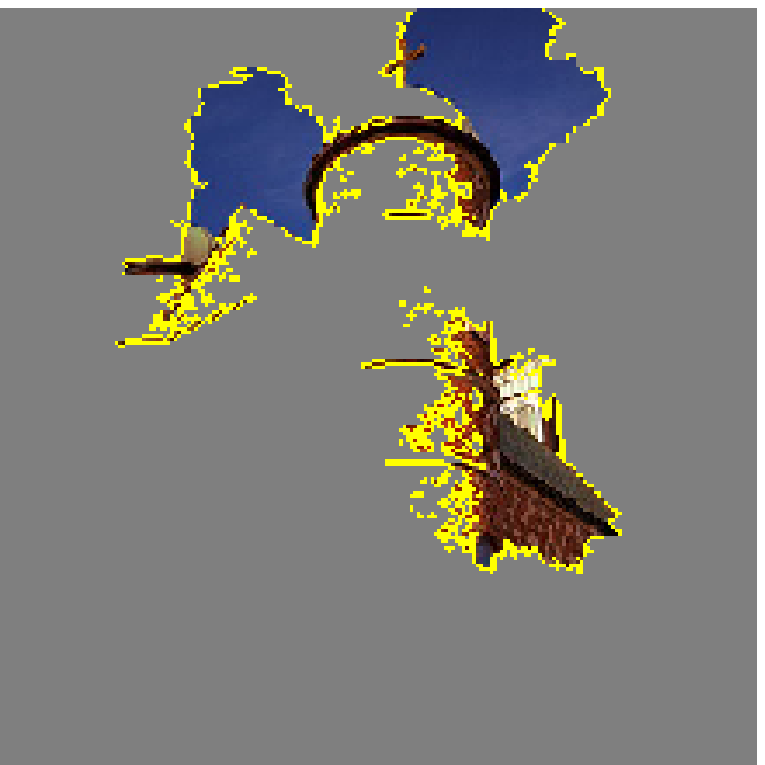}}
  \centerline{\scriptsize{(d) LIME (K=4)}}
  %\centerline{}
\end{minipage}
\vfill
\begin{minipage}{0.24\linewidth}
  %\centering
  \centerline{\includegraphics[width=1.0\textwidth]{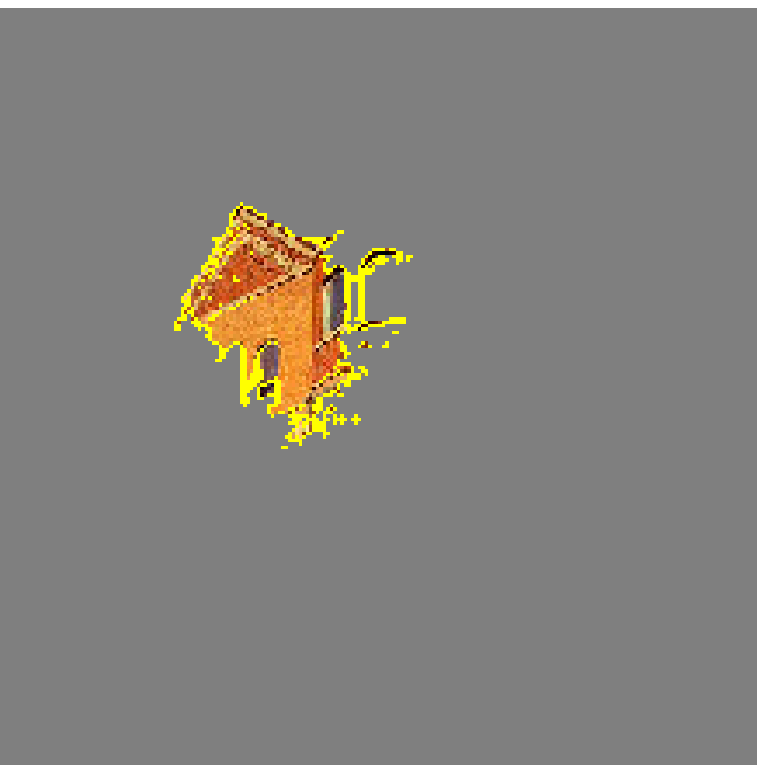}}
  \centerline{\scriptsize{(a) LEGKC (K=1)}}
  %\centerline{}
\end{minipage}
\hfill
\begin{minipage}{0.24\linewidth}
  %\centering
  \centerline{\includegraphics[width=1.0\textwidth]{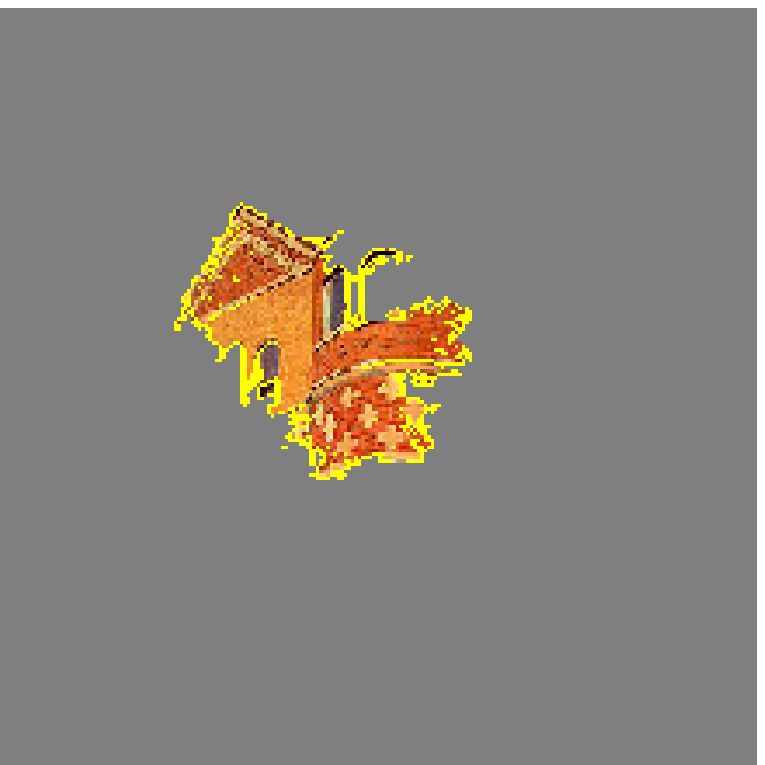}}
  \centerline{\scriptsize{(b) LEGKC (K=2)}}
  %\centerline{}
\end{minipage}
\hfill
\begin{minipage}{0.24\linewidth}
  %\centering
  \centerline{\includegraphics[width=1.0\textwidth]{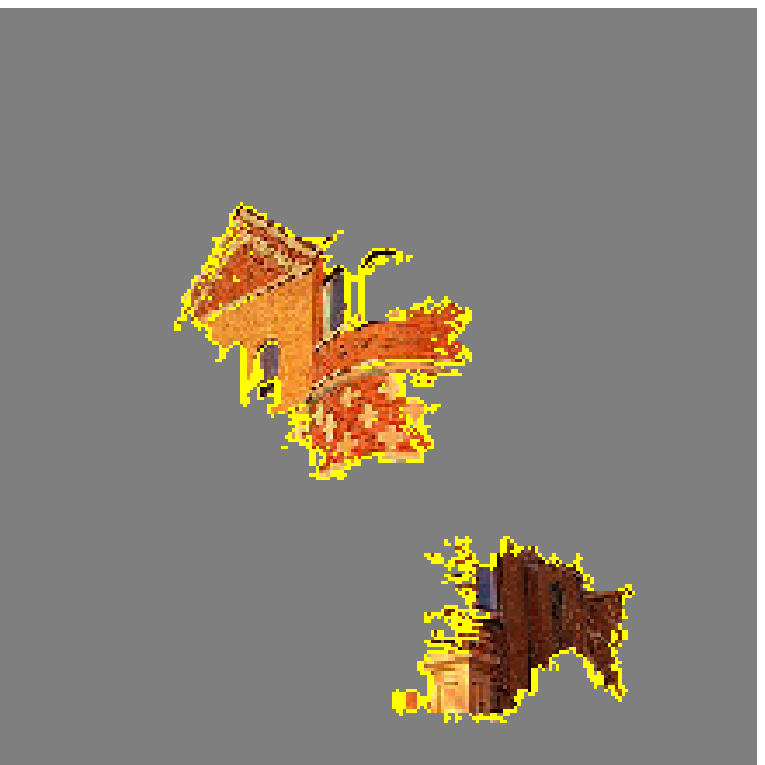}}
  \centerline{\scriptsize{(c) LEGKC (K=3)}}
  %\centerline{}
\end{minipage}
\hfill
\begin{minipage}{0.24\linewidth}
  %\centering
  \centerline{\includegraphics[width=1.0\textwidth]{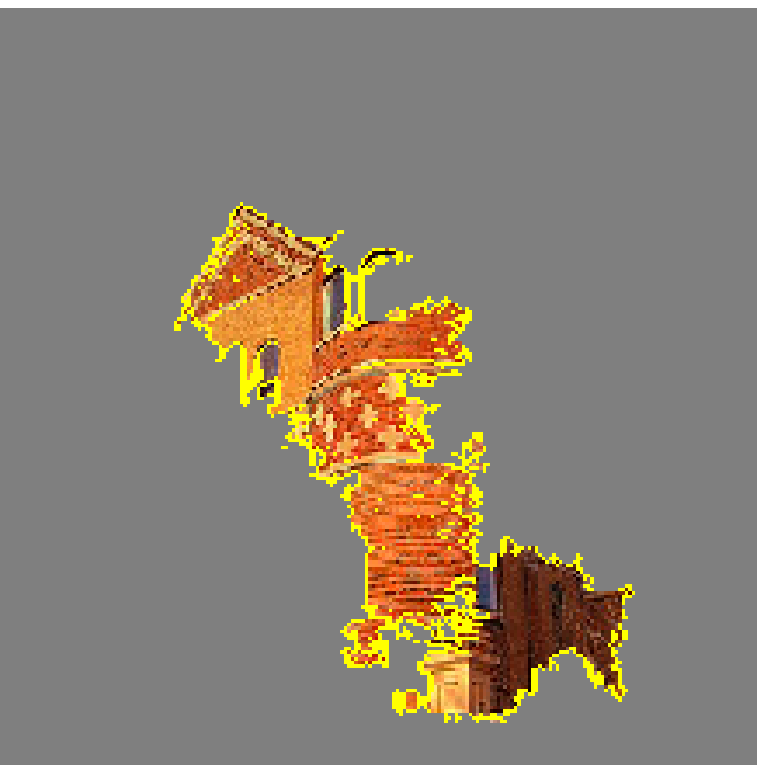}}
  \centerline{\scriptsize{(d) LEGKC (K=4)}}
  %\centerline{}
\end{minipage}
\caption{\small{Explaining image classification predictions made by Google's Inception neural network. The first row shows the superpixels explanations by LIME. The second row shows the superpixels explanations by LEGKC.}}
\label{fig:4}
\end{figure}

In this section, LEDSNA and LIME explain image classification predictions made by Google's pre-trained Inception neural network \cite{Inception}. Fig.~\ref{fig:3} shows two original image to be processed. Fig.~\ref{fig:4} and Fig.~\ref{fig:5} lists some visual explanations of LEDSNA and LIME: the first row shows the superpixels explanations by LIME (K=1,2,3,4) respectively, the second row shows the superpixels explanations by LEDSNA (K=1,2,3,4) respectively. The explanations highlight the top K superpixel segments, which have the most considerable positive weights towards the predictions. We can see LEDSNA can effectively get the correlation between the adjacent superpixel segments, which provide a better understanding to users.
\begin{figure}
\begin{minipage}{0.24\linewidth}
  %\centering
  \centerline{\includegraphics[width=1.0\textwidth]{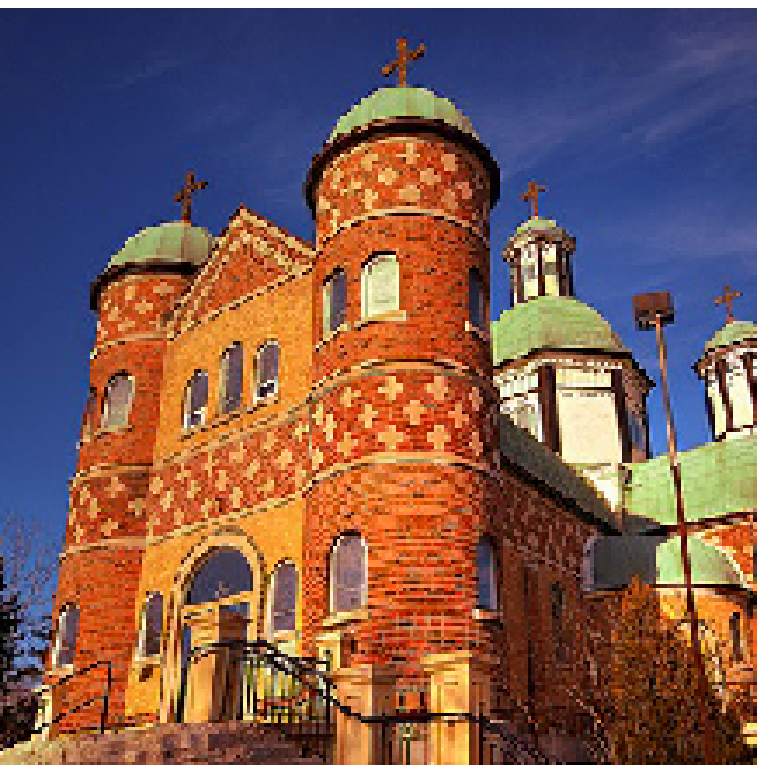}}
  %\centerline{\scriptsize{(a)}}
  %\centerline{}
\end{minipage}
\hfill
\begin{minipage}{0.24\linewidth}
  %\centering
  \centerline{\includegraphics[width=1.0\textwidth]{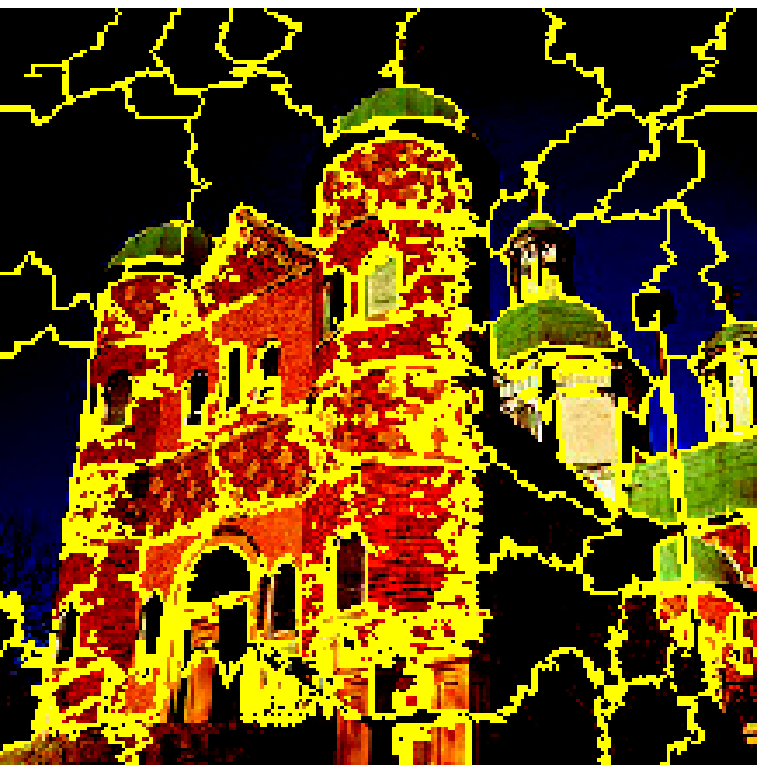}}
  %\centerline{\scriptsize{(a)}}
  %\centerline{}
\end{minipage}
\hfill
\begin{minipage}{0.24\linewidth}
  %\centering
  \centerline{\includegraphics[width=1.0\textwidth]{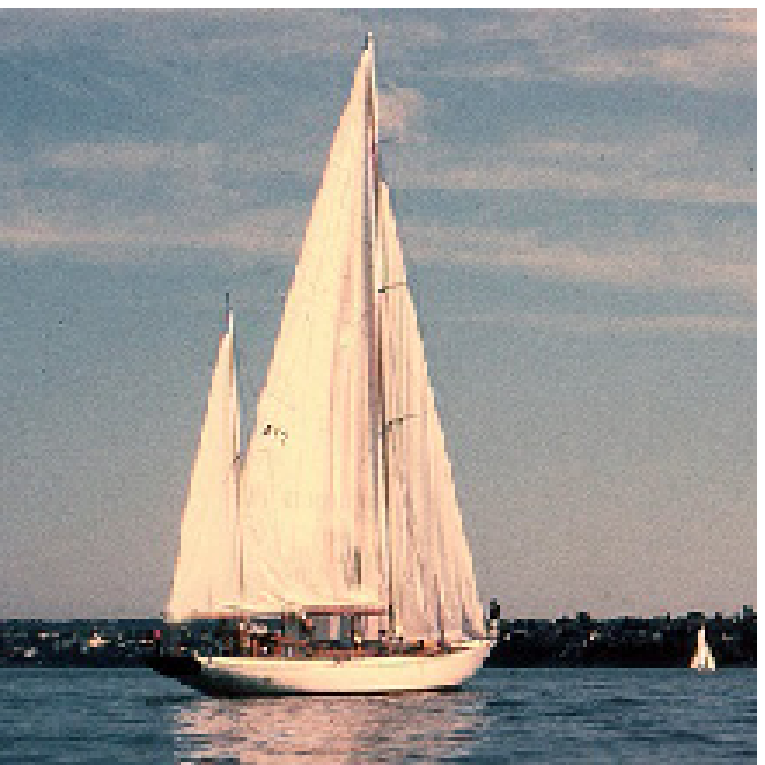}}
  %\centerline{\scriptsize{(b)}}
  %\centerline{}
\end{minipage}
\hfill
\begin{minipage}{0.24\linewidth}
  %\centering
  \centerline{\includegraphics[width=1.0\textwidth]{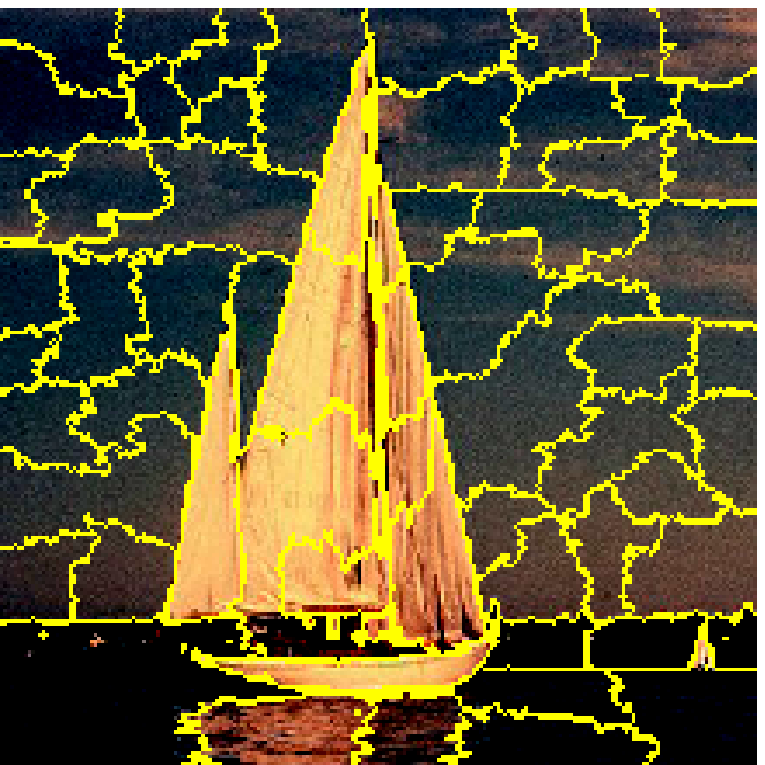}}
  %\centerline{\scriptsize{(b)}}
  %\centerline{}
\end{minipage}
\caption{\small{Original images and superpixel images}}
\label{fig:3}
\end{figure}

\begin{figure}
\begin{minipage}{0.24\linewidth}
  %\centering
  \centerline{\includegraphics[width=1.0\textwidth]{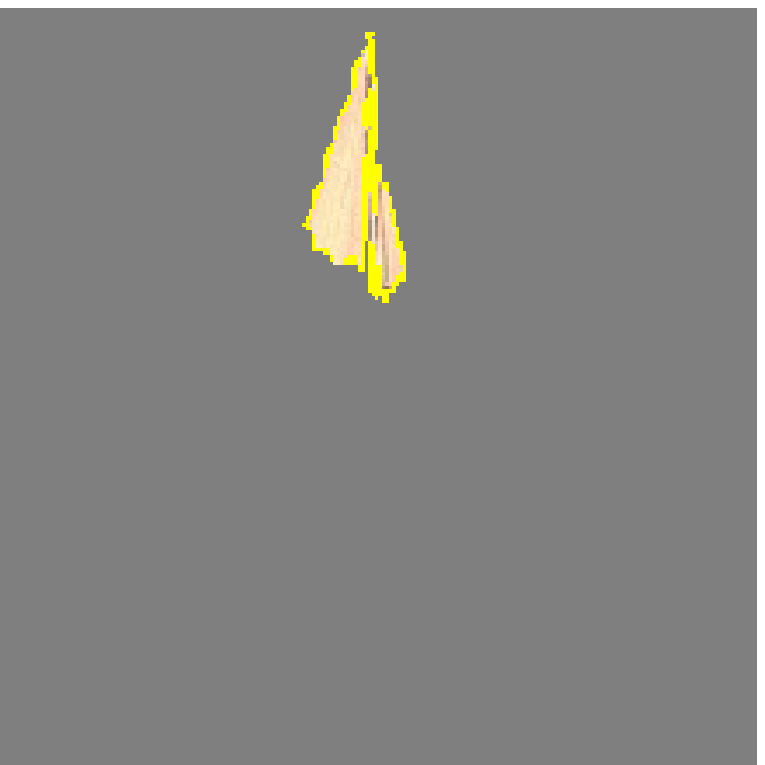}}
  \centerline{\scriptsize{(a) LIME (K=1)}}
  %\centerline{}
\end{minipage}
\hfill
\begin{minipage}{0.24\linewidth}
  %\centering
  \centerline{\includegraphics[width=1.0\textwidth]{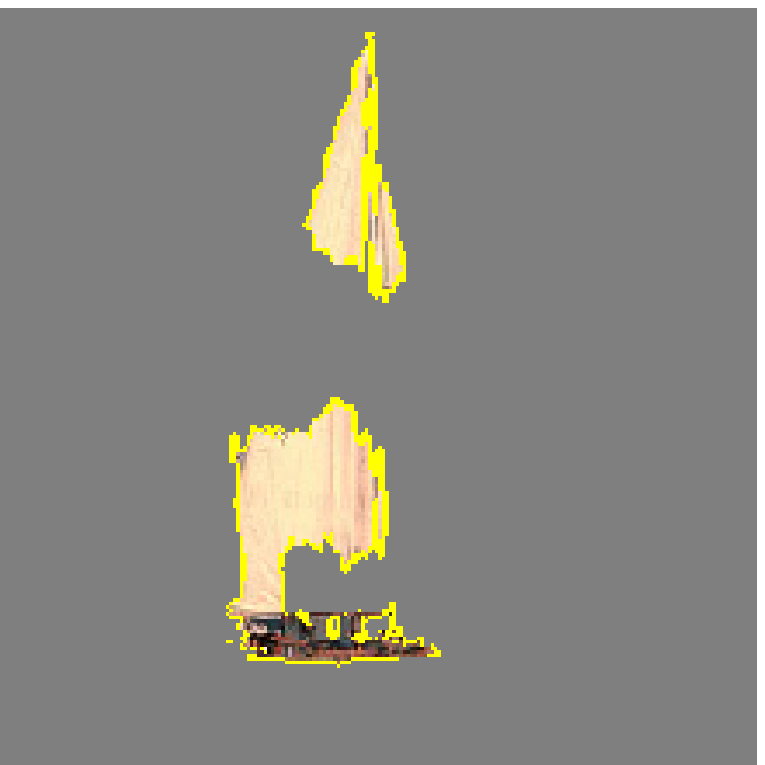}}
  \centerline{\scriptsize{(b) LIME (K=2)}}
  %\centerline{}
\end{minipage}
\hfill
\begin{minipage}{0.24\linewidth}
  %\centering
  \centerline{\includegraphics[width=1.0\textwidth]{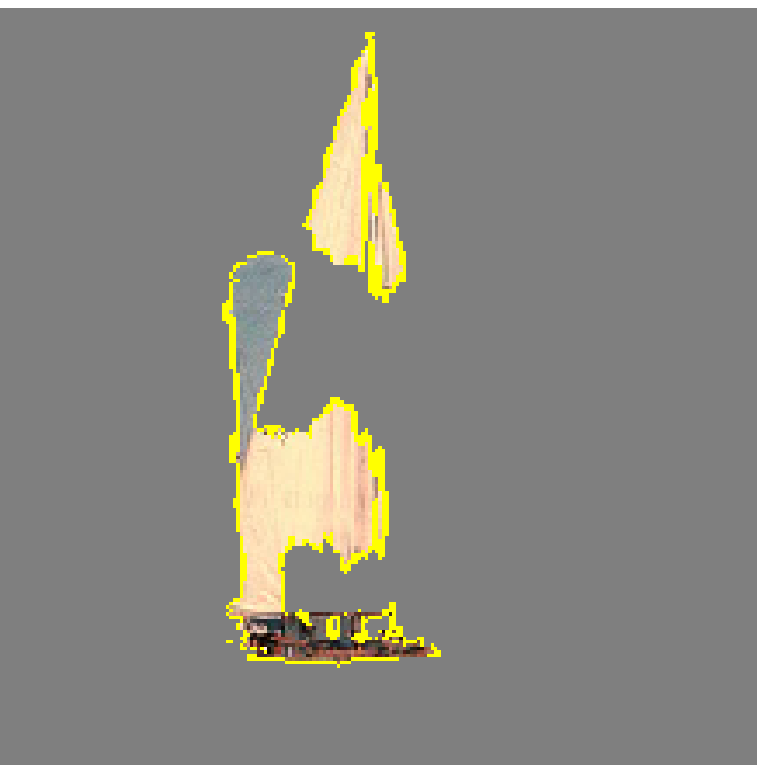}}
  \centerline{\scriptsize{(c) LIME (K=3)}}
  %\centerline{}
\end{minipage}
\hfill
\begin{minipage}{0.24\linewidth}
  %\centering
  \centerline{\includegraphics[width=1.0\textwidth]{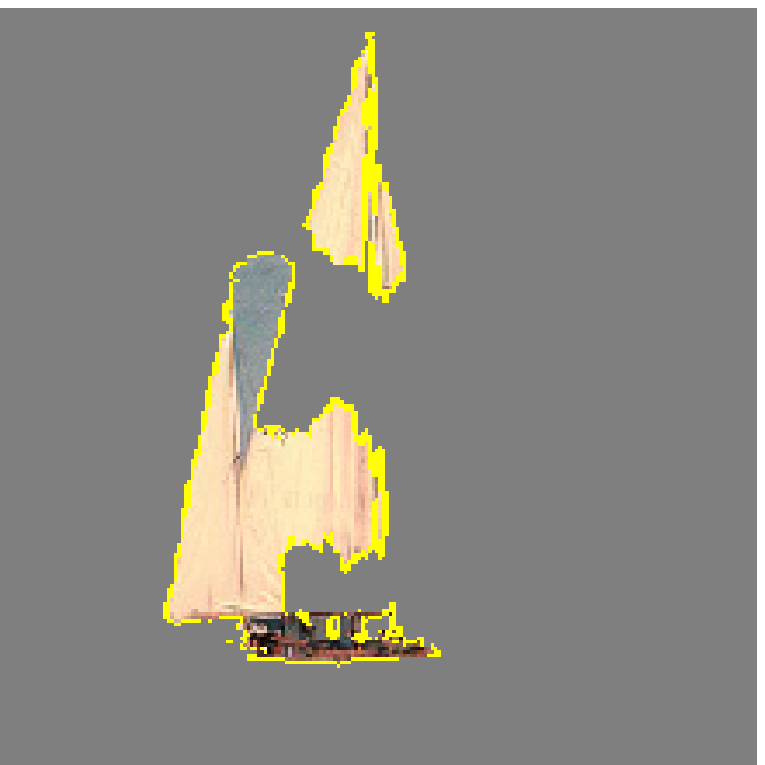}}
  \centerline{\scriptsize{(d) LIME (K=4)}}
  %\centerline{}
\end{minipage}
\vfill
\begin{minipage}{0.24\linewidth}
  %\centering
  \centerline{\includegraphics[width=1.0\textwidth]{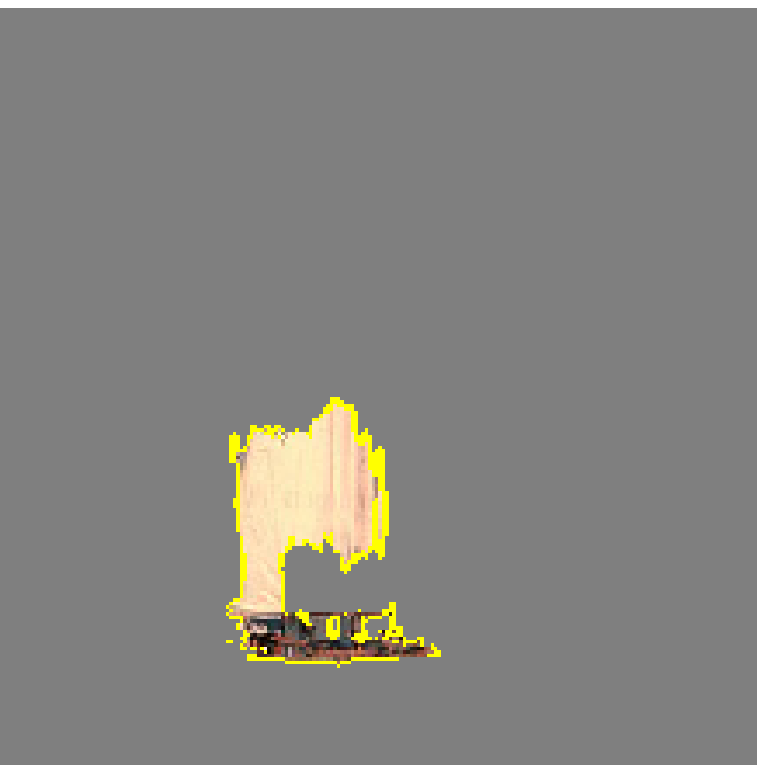}}
  \centerline{\scriptsize{(a) LEGKC (K=1)}}
  %\centerline{}
\end{minipage}
\hfill
\begin{minipage}{0.24\linewidth}
  %\centering
  \centerline{\includegraphics[width=1.0\textwidth]{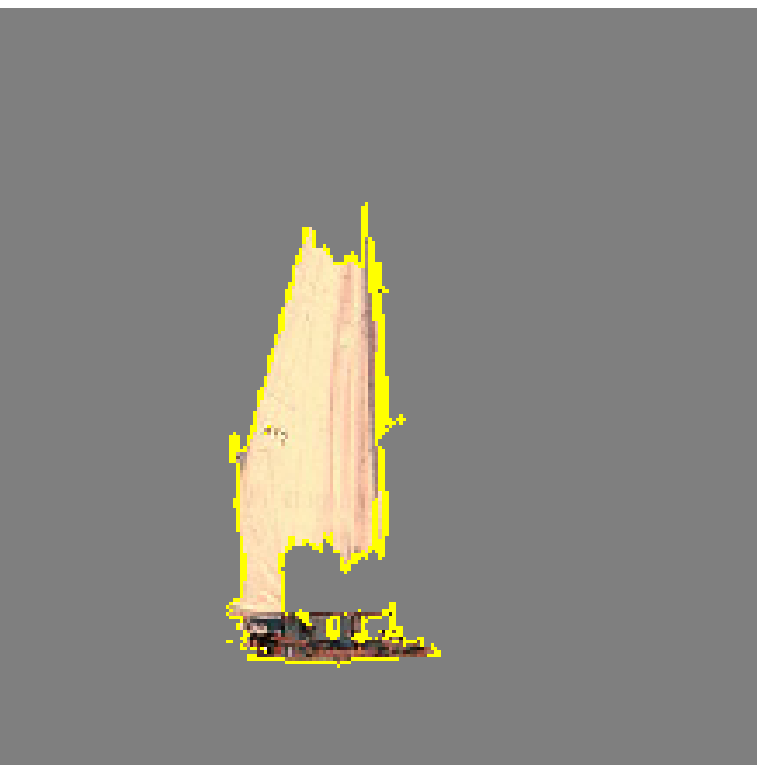}}
  \centerline{\scriptsize{(b) LEGKC (K=2)}}
  %\centerline{}
\end{minipage}
\hfill
\begin{minipage}{0.24\linewidth}
  %\centering
  \centerline{\includegraphics[width=1.0\textwidth]{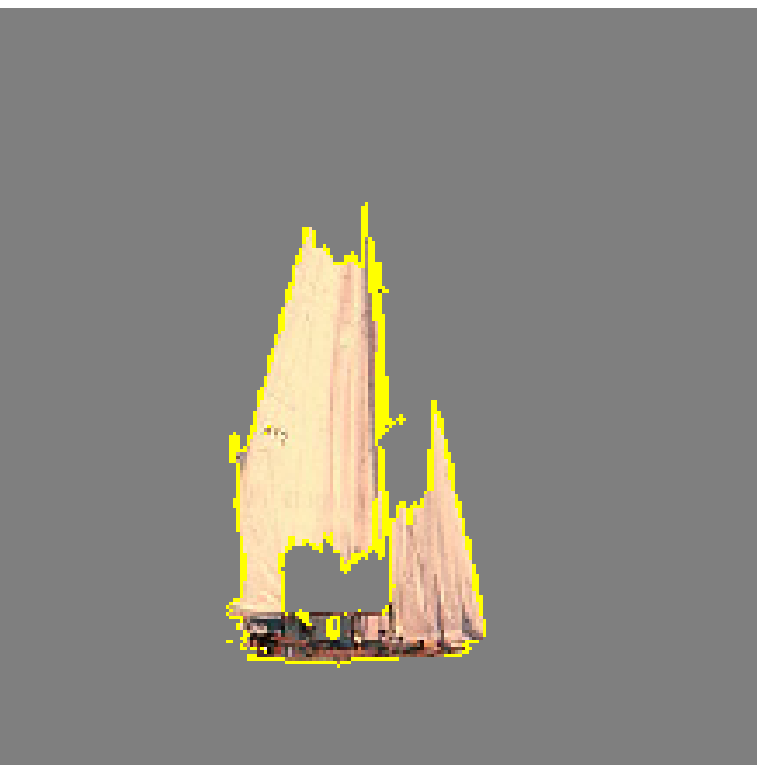}}
  \centerline{\scriptsize{(c) LEGKC (K=3)}}
  %\centerline{}
\end{minipage}
\hfill
\begin{minipage}{0.24\linewidth}
  %\centering
  \centerline{\includegraphics[width=1.0\textwidth]{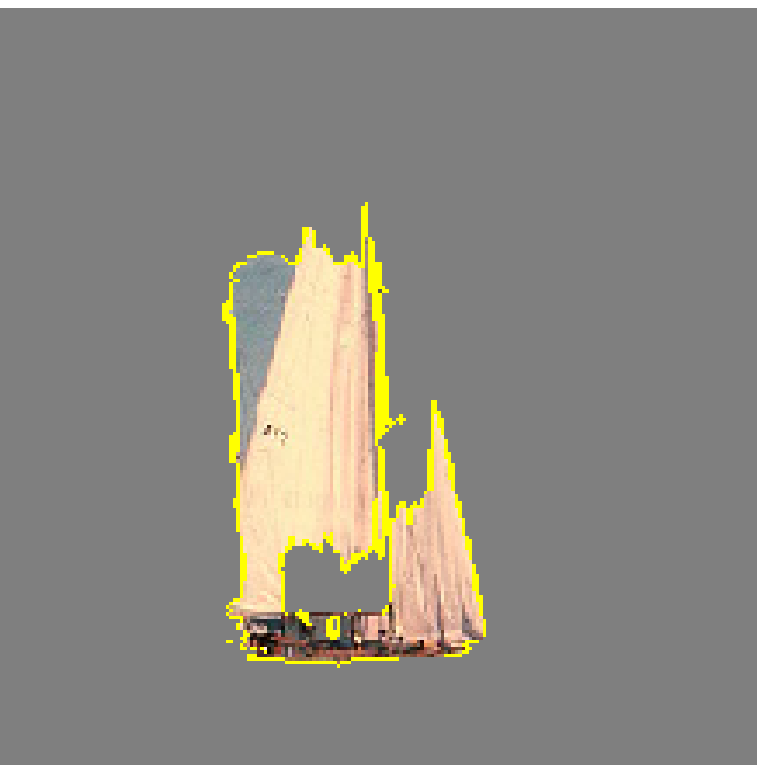}}
  \centerline{\scriptsize{(d) LEGKC (K=4)}}
  %\centerline{}
\end{minipage}
\caption{\small{Explaining image classification predictions made by Google's Inception neural network. The first row shows the superpixels explanations by LIME. The second row shows the superpixels explanations by LEGKC.}}
\label{fig:5}
\end{figure}

In addition, Table~\ref{tab:1} lists some instances of the local approximation error and $R^2$ of two algorithm. Comparing to LIME, we can see LEDSNA provides better predictive accuracy than LIME. Besides, $R^2$ of LEDSNA is much bigger than LIME. By comparing the two criterion, we conclude that LEDSNA has better fidelity than LIME. Compared with LIME in term of interpretability and fidelity, LEDSNA has better performance in explaining classification.

\begin{table}
  \centering\footnotesize
  \caption{\small{Comparison of LIME and LEDSNA in the task of image classification}}
    \begin{tabular}{|c|c|c|c|c|}
    %\begin{tabular}{|c|c|c|c|}
    \hline
     & f(x) & g(x)  & $Err$ & $R^2$ \bigstrut\\
    \hline
    LIME &\multirow{2}{*}{$p_{yawl}=0.6076$}&0.8129 &0.2053 &0.4662  \bigstrut\\
    %\hline
    LEDSNA  & & 0.6066 & 0.001 &0.9803 \bigstrut\\
    \hline
    LIME &\multirow{2}{*}{$p_{castle}=0.7646$}& 0.9857 & 0.2211 &0.3219 \bigstrut\\
    %\hline
    LEDSNA  & & 0.7633 & 0.0012 &0.896\bigstrut\\
    \hline
    LIME &\multirow{2}{*}{$p_{church}=0.2885$}& 0.5133 & 0.2248 &0.4644 \bigstrut\\
    %\hline
    LEDSNA  & & 0.288 & 0.0005 &0.5890\bigstrut\\
    \hline
    LIME &\multirow{2}{*}{$p_{butterfly}=0.9035$}& 1.5995 &0.6194 &0.5939 \bigstrut\\
    %\hline
    LEDSNA  & & 0.9025 & 0.0010 &0.8407\bigstrut\\
    \hline
    LIME &\multirow{2}{*}{$p_{magpie}=0.9461$}& 1.2854 & 0.2655 &0.3602 \bigstrut\\
    %\hline
    LEDSNA  & & 0.945 & 0.0010 & 0.7955\bigstrut\\
    \hline
    LIME &\multirow{2}{*}{$p_{liner}=0.9669$}& 1.2422 & 0.2753  &0.6341 \bigstrut\\
    %\hline
    LEDSNA  & &  0.9657 & 0.0012 &0.8414 \bigstrut\\
    \hline
    \end{tabular}
    %\vspace{-50pt}
  \label{tab:1}
\end{table}

\subsection{Experiment on Sentiment Analysis of Text}

\subsubsection{Experiment on Chinese Natural Language Databse}
Simplified Chinese Text Processing (SnowNLP) is a sentiment analysis tool especially for Chinese natural language. This section we use LEDSNA and LIME to explain the predictions made by SnowNLP on Public Comment Dataset. As there is a strong semantic dependency between words in Chinese, we incorporate the Stanford Word Segmenter \cite{CoreNLP} into sampling process to get the perturbed samples. In nonlinear approximating, we use Gaussian kernel function to compute the similarity between the data points in a much higher dimensional space.
% \begin{equation}
% k(x,x')=e^{-(x-x')^2/\sigma^2}.
% \end{equation}
\begin{figure}
\centering
\begin{minipage}{0.8\linewidth}
  %\centering
  \centerline{\includegraphics[width=1.0\textwidth]{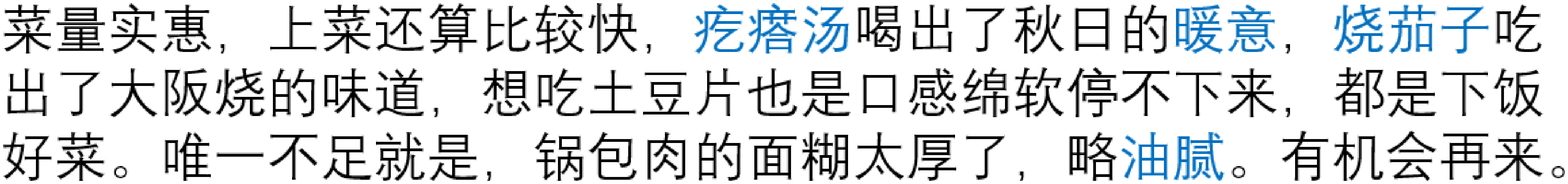}}
  \centerline{\scriptsize{(a) LEGKC}}
  %\centerline{}
\end{minipage}
\hfill
\begin{minipage}{0.8\linewidth}
  %\centering
  \centerline{\includegraphics[width=1.0\textwidth]{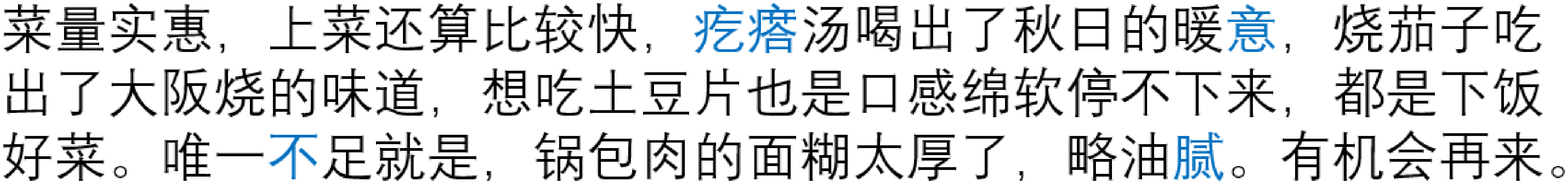}}
  \centerline{\scriptsize{(b) LIME}}
  %\centerline{}
\end{minipage}
\caption{\small{Sentiment analysis (p=0.9843)}}
\label{fig:6}
\end{figure}

\begin{figure}
\centering
\begin{minipage}{0.6\linewidth}
  %\centering
  \centerline{\includegraphics[width=1.0\textwidth]{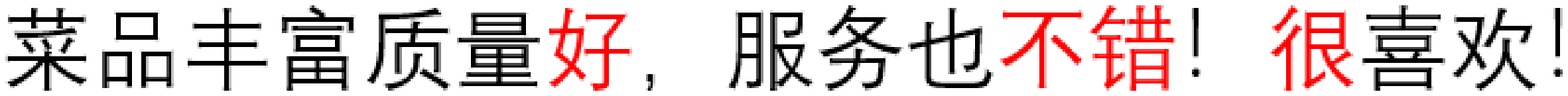}}
  \centerline{\scriptsize{(a) LEGKC}}
  %\centerline{}
\end{minipage}
\hfill
\begin{minipage}{0.6\linewidth}
  %\centering
  \centerline{\includegraphics[width=1.0\textwidth]{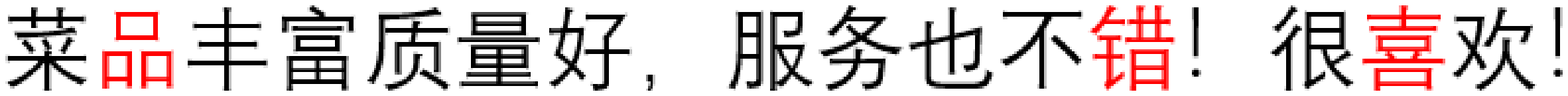}}
  \centerline{\scriptsize{(b) LIME}}
  %\centerline{}
\end{minipage}
\caption{\small{Sentiment analysis (p=0.022)}}
\label{fig:7}
\end{figure}

Fig.~\ref{fig:6} and Fig.~\ref{fig:7} shows visual explanations of LEDNSA and LIME, we can see the explanation of LEDNSA can offer more useful information than that of LIME. Table~\ref{tab:2} lists the local approximation error and $R^2$ of six instances. Comparing to LIME, we find LEDSNA achieves better performance across the board, and by average a magnitude of local approximation error than LIME. For $R^2$, similar observation is obtained.

\begin{table}
  \centering\footnotesize
  \caption{\small{Comparison of LIME and LEDSNA in the task of text classification}}
    \begin{tabular}{|c|c|c|c|c|}
    %\begin{tabular}{|c|c|c|c|}
    \hline
     & f(x) & g(x) & $R^2$ & $Err$ \bigstrut\\
    \hline
    LIME &\multirow{2}{*}{0.1788} & 0.1755 & 0.4795 &0.0033  \bigstrut\\
    %\hline
    LEDSNA  & & 0.1765 & 0.9973 & 0.0023  \bigstrut\\
    \hline
    LIME &\multirow{2}{*}{0.1224} & 0.1136 & 0.4969  &0.0088 \bigstrut\\
    %\hline
    LEDSNA  & & 0.1209 & 0.8710 & 0.0016 \bigstrut\\
    \hline
    LIME &\multirow{2}{*}{0.2298} & 0.3082& 0.4823 & 0.0784  \bigstrut\\
    %\hline
    LEDSNA  & & 0.2283 & 0.9790 & 0.0015\bigstrut\\
    \hline
    LIME &\multirow{2}{*}{0.4839} & 0.3526 & 0.5876 &0.1313 \bigstrut\\
    %\hline
    LEDSNA  & & 0.4756 & 0.9822 & 0.0083\bigstrut\\
    \hline
    LIME &\multirow{2}{*}{0.6489} & 0.6901& 0.4449 & 0.0419 \bigstrut\\
    %\hline
    LEDSNA  & & 0.6482 & 0.9473 & 0.0008\bigstrut\\
    \hline
    LIME &\multirow{2}{*}{0.9052} & 0.8717 & 0.5779 & 0.0335 \bigstrut\\
    %\hline
    LEDSNA  & & 0.9050 & 0.9533 & 0.0001\bigstrut\\
    \hline
    \end{tabular}
    %\vspace{-50pt}
  \label{tab:2}
\end{table}

\begin{figure}
\centering
\begin{minipage}{0.8\linewidth}
  \centering
  \centerline{\includegraphics[width=1.0\textwidth]{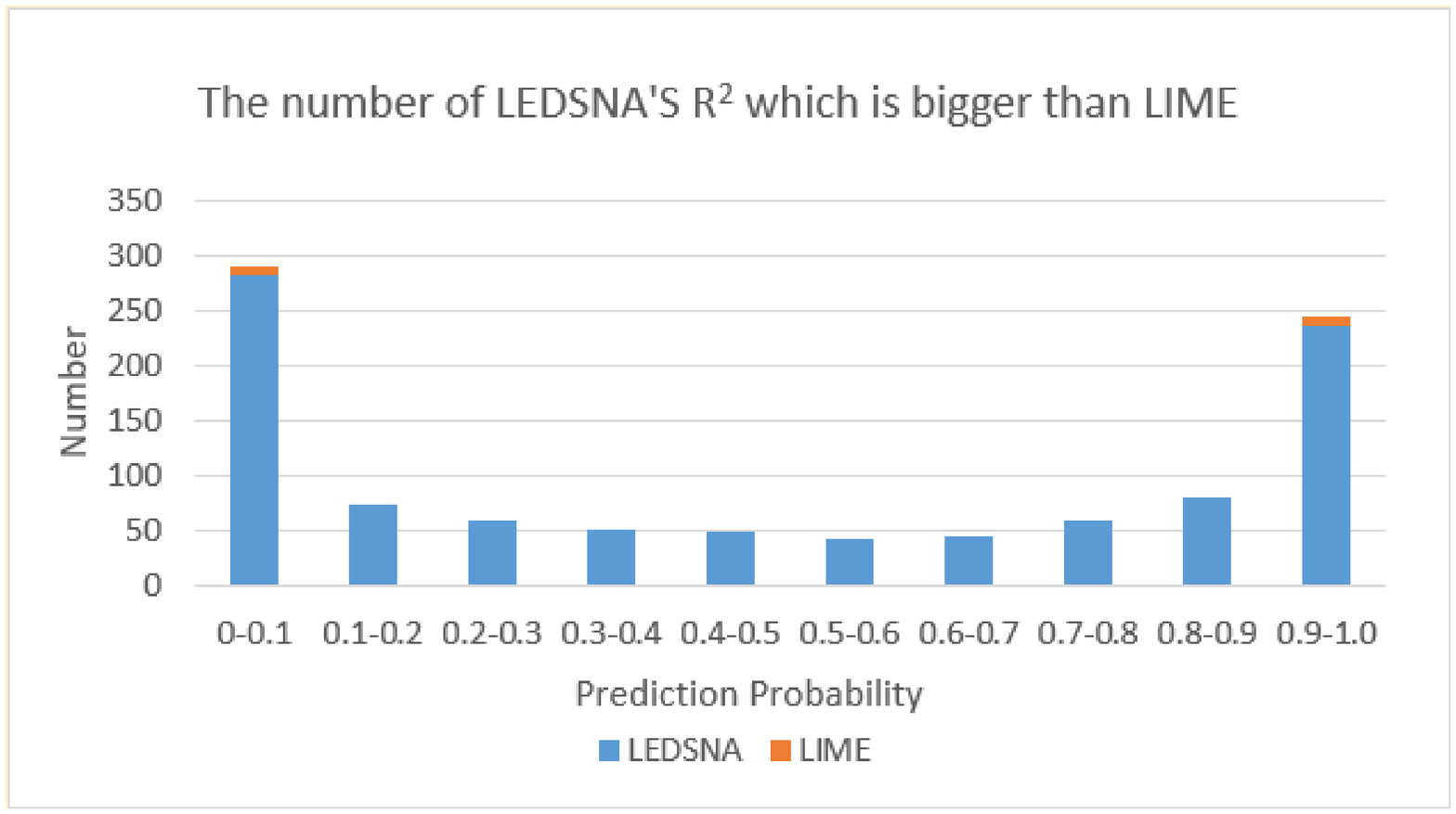}}
  \centerline{\scriptsize{(a)}}
  %\centerline{}
\end{minipage}
\vfill
\begin{minipage}{0.8\linewidth}
  \centering
  \centerline{\includegraphics[width=1.0\textwidth]{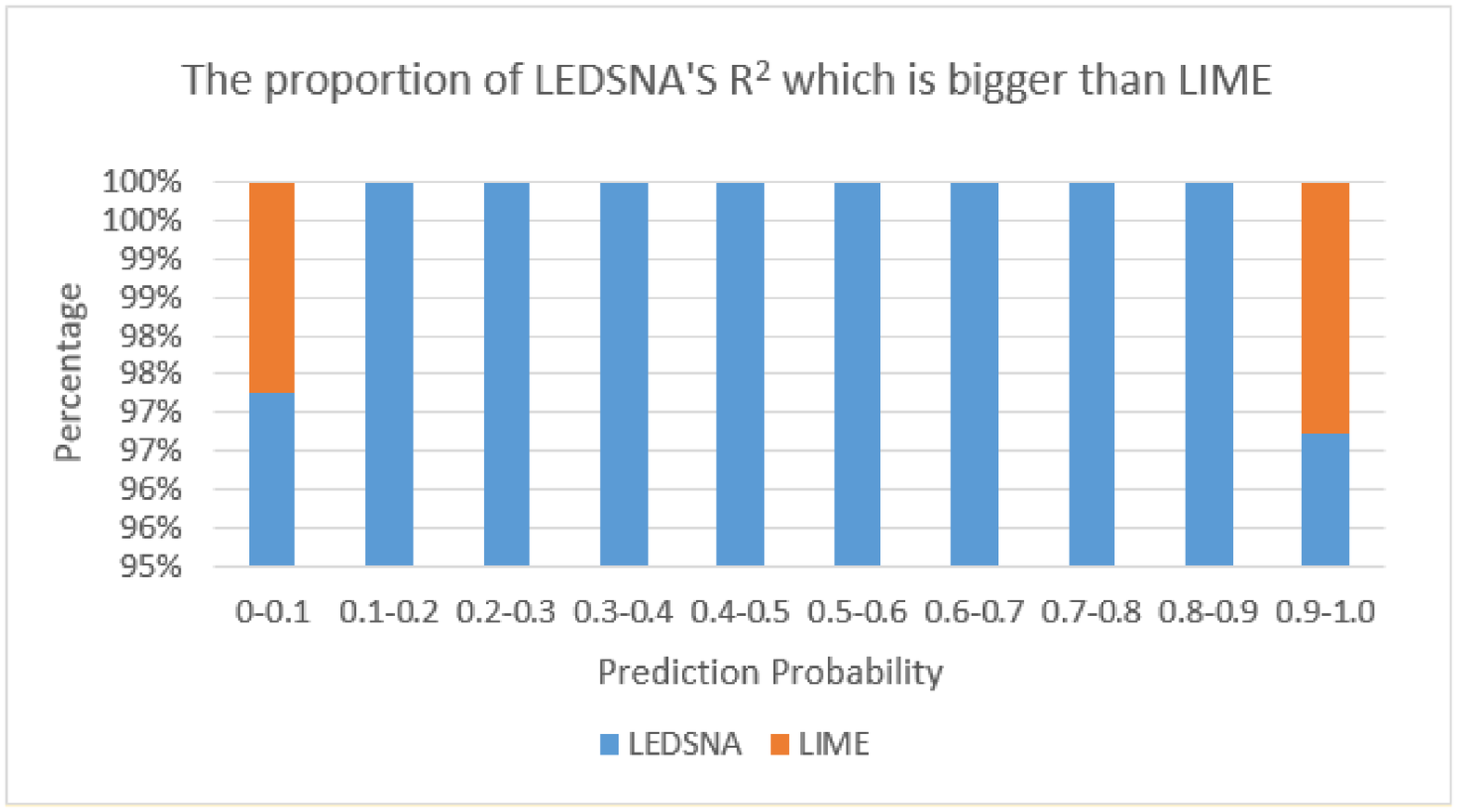}}
  \centerline{\scriptsize{(b)}}
  %\centerline{}
\end{minipage}
\caption{\small{The $R^2$ of $1000$ test cases: (a) The number of LEDSNA's $R^2$ which is bigger than LIME. (b) The proportion of LEDSNA's $R^2$ which is bigger than LIME.}}
\label{fig:8}
\end{figure}

\begin{figure}
\centering
\begin{minipage}{0.8\linewidth}
  \centering
  \centerline{\includegraphics[width=1.0\textwidth]{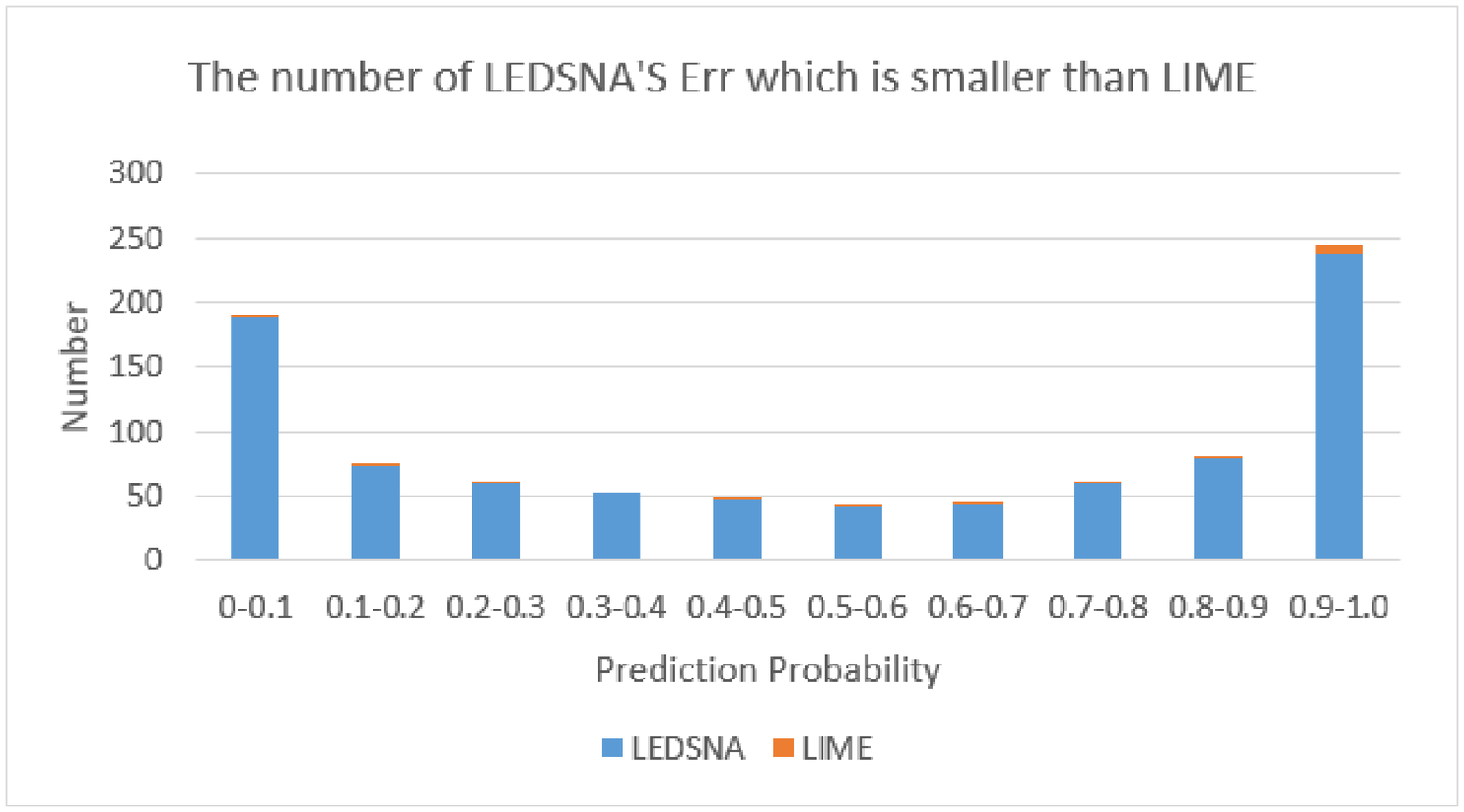}}
  \centerline{\scriptsize{(a)}}
  %\centerline{}
\end{minipage}
\vfill
\begin{minipage}{0.8\linewidth}
  \centering
  \centerline{\includegraphics[width=1.0\textwidth]{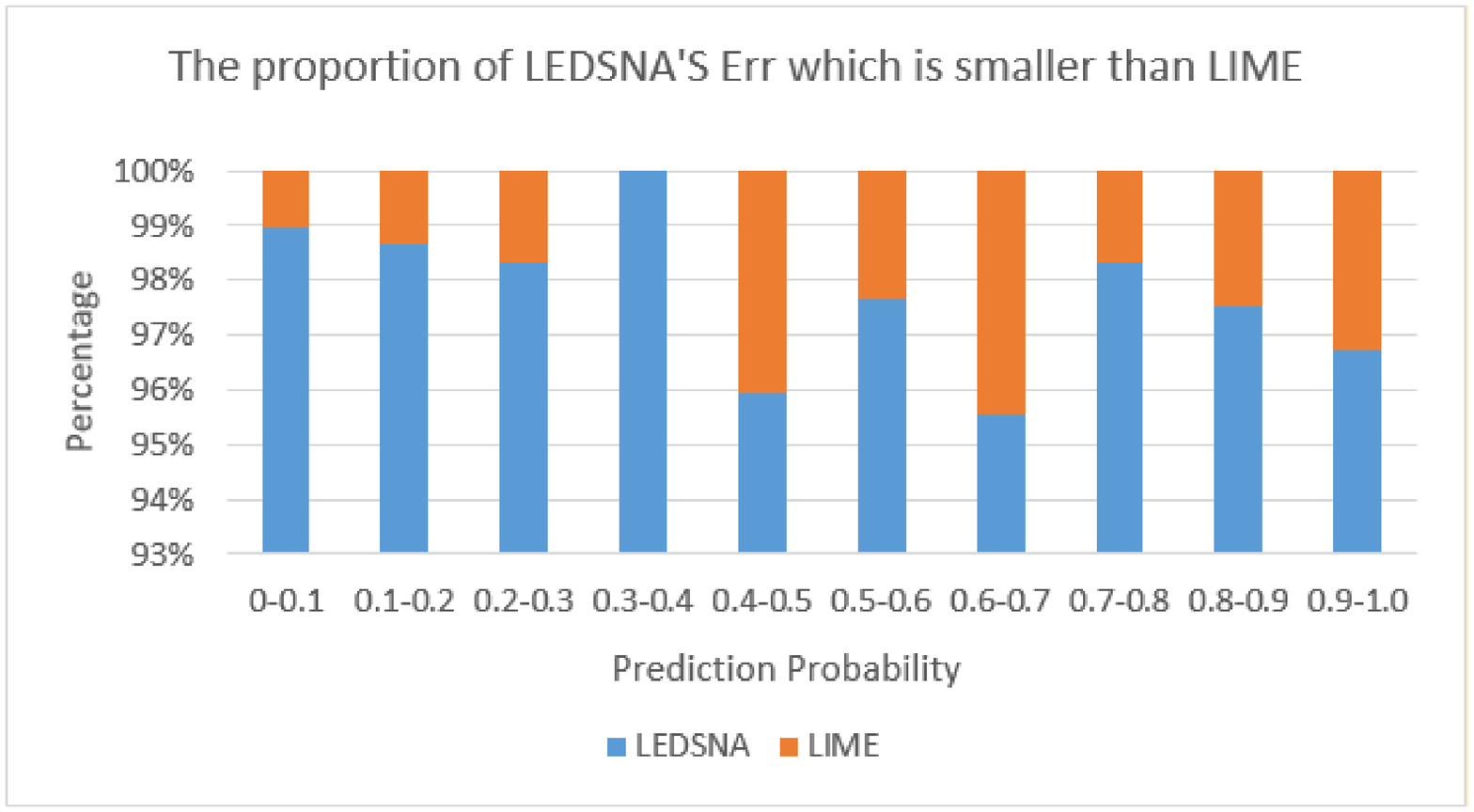}}
  \centerline{\scriptsize{(b)}}
  %\centerline{}
\end{minipage}
\caption{\small{The Err of $1000$ test cases: (a) The number of LEDSNA's Err which is smaller than LIME. (b) The proportion of LEDSNA's Err which is smaller than LIME.}}
\label{fig:9}
\end{figure}

Moreover, we randomly selected 1000 data samples to constitute testing database. For each testing data sample, we use LEDSNA and LIME to explain SnowNLP and compute the $Err$ and $R^2$. Results show for LEDSNA, the Err of $95\%$ of test data samples are smaller than LIME. Similarly, the $R^2$ of $98.4\%$ of test data samples are bigger than LIME. In conclusion, LEDSNA exhibits strong interpretability and fidelity over LIME

\section{Conclusion}
There are two drawbacks in current existing local explanations. Perturbed samples are generated from a uniform distribution, ignoring the complicated correlation between features. This may lead to lose much useful information to learn the local explanation models. Moreover, most existing methods assume the decision boundary is local linearity, which may produce serious errors as in most complex networks, the local decision boundary is non-linear.

In this paper, we design and develop a novel, high-fidelity local explanation method to address the above challenges. First, we design a unique local sampling process which incorporate the feature clustering method to handle the feature dependency problems. Then, we adopt SVR to approximate locally nonlinear boundary. In this way, by simultaneously preserving feature dependency and local non-linearity, our method produces high-fidelity and high-interpretability explanation.

\newpage
\bibliographystyle{IEEEtran}
\bibliography{refs}

\begin{thebibliography}{10}
\providecommand{\url}[1]{#1}
\csname url@samestyle\endcsname
\providecommand{\newblock}{\relax}
\providecommand{\bibinfo}[2]{#2}
\providecommand{\BIBentrySTDinterwordspacing}{\spaceskip=0pt\relax}
\providecommand{\BIBentryALTinterwordstretchfactor}{4}
\providecommand{\BIBentryALTinterwordspacing}{\spaceskip=\fontdimen2\font plus
\BIBentryALTinterwordstretchfactor\fontdimen3\font minus
  \fontdimen4\font\relax}
\providecommand{\BIBforeignlanguage}[2]{{%
\expandafter\ifx\csname l@#1\endcsname\relax
\typeout{** WARNING: IEEEtran.bst: No hyphenation pattern has been}%
\typeout{** loaded for the language `#1'. Using the pattern for}%
\typeout{** the default language instead.}%
\else
\language=\csname l@#1\endcsname
\fi
#2}}
\providecommand{\BIBdecl}{\relax}
\BIBdecl

\bibitem{AI01}
I.~Goodfellow, Y.~Bengio, and A.~Courville, \emph{Deep Learning}.\hskip 1em
  plus 0.5em minus 0.4em\relax MIT Press, 2016,
  \url{http://www.deeplearningbook.org}.

\bibitem{AI02}
T.~Hastie, R.~Tibshirani, and J.~Friedman, \emph{The Elements of Statistical
  Learning}.\hskip 1em plus 0.5em minus 0.4em\relax Springer, 2009,
  \url{www.web.stanford.edu/~hastie/ElemStatLearn}.

\bibitem{AI03}
S.~{Ren}, K.~{He}, R.~{Girshick}, and J.~{Sun}, ``Faster r-cnn: Towards
  real-time object detection with region proposal networks,'' \emph{IEEE
  Transactions on Pattern Analysis and Machine Intelligence}, vol.~39, no.~6,
  pp. 1137--1149, June 2017.

\bibitem{EX01}
\BIBentryALTinterwordspacing
Y.~Lou, R.~Caruana, and J.~Gehrke, ``Intelligible models for classification and
  regression,'' in \emph{The 18th {ACM} {SIGKDD} International Conference on
  Knowledge Discovery and Data Mining, {KDD} '12, Beijing, China, August 12-16,
  2012}, 2012, pp. 150--158. [Online]. Available:
  \url{https://doi.org/10.1145/2339530.2339556}
\BIBentrySTDinterwordspacing

\bibitem{EX02}


\bibitem{EX03}
\BIBentryALTinterwordspacing
C.~Szegedy, W.~Zaremba, I.~Sutskever, J.~Bruna, D.~Erhan, I.~J. Goodfellow, and
  R.~Fergus, ``Intriguing properties of neural networks,'' in \emph{2nd
  International Conference on Learning Representations, {ICLR} 2014, Banff, AB,
  Canada, April 14-16, 2014, Conference Track Proceedings}, 2014. [Online].
  Available: \url{http://arxiv.org/abs/1312.6199}
\BIBentrySTDinterwordspacing

\bibitem{POST01}
\BIBentryALTinterwordspacing
M.~T. Ribeiro, S.~Singh, and C.~Guestrin, ``Why should {I} trust you?:
  Explaining the predictions of any classifier,'' in \emph{Proceedings of the
  22nd {ACM} {SIGKDD} International Conference on Knowledge Discovery and Data
  Mining, San Francisco, CA, USA, August 13-17, 2016}, 2016, pp. 1135--1144.
  [Online]. Available: \url{https://doi.org/10.1145/2939672.2939778}
\BIBentrySTDinterwordspacing

\bibitem{POST02}
I.~Guyon, U.~von Luxburg, S.~Bengio, H.~M. Wallach, R.~Fergus, S.~V.~N.
  Vishwanathan, and R.~Garnett, Eds., \emph{Advances in Neural Information
  Processing Systems 30: Annual Conference on Neural Information Processing
  Systems 2017, 4-9 December 2017, Long Beach, CA, {USA}}, 2017.

\bibitem{POST03}
\BIBentryALTinterwordspacing
M.~T. Ribeiro, S.~Singh, and C.~Guestrin, ``Model-agnostic interpretability of
  machine learning,'' \emph{CoRR}, vol. abs/1606.05386, 2016. [Online].
  Available: \url{http://arxiv.org/abs/1606.05386}
\BIBentrySTDinterwordspacing

\bibitem{App01}
\BIBentryALTinterwordspacing
S.~Tan, R.~Caruana, G.~Hooker, and Y.~Lou, ``Distill-and-compare: Auditing
  black-box models using transparent model distillation,'' in \emph{Proceedings
  of the 2018 {AAAI/ACM} Conference on AI, Ethics, and Society, {AIES} 2018,
  New Orleans, LA, USA, February 02-03, 2018}, 2018, pp. 303--310. [Online].
  Available: \url{https://doi.org/10.1145/3278721.3278725}
\BIBentrySTDinterwordspacing

\bibitem{App02}
\BIBentryALTinterwordspacing
R.~E. Shawi, M.~H. Al{-}Mallah, and S.~Sakr, ``On the interpretability of
  machine learning-based model for predicting hypertension,'' \emph{{BMC} Med.
  Inf. {\&} Decision Making}, vol.~19, no.~1, pp. 146:1--146:32, 2019.
  [Online]. Available: \url{https://doi.org/10.1186/s12911-019-0874-0}
\BIBentrySTDinterwordspacing

\bibitem{Other01}
\BIBentryALTinterwordspacing
H.~Lakkaraju, S.~H. Bach, and J.~Leskovec, ``Interpretable decision sets: {A}
  joint framework for description and prediction,'' in \emph{Proceedings of the
  22nd {ACM} {SIGKDD} International Conference on Knowledge Discovery and Data
  Mining, San Francisco, CA, USA, August 13-17, 2016}, 2016, pp. 1675--1684.
  [Online]. Available: \url{https://doi.org/10.1145/2939672.2939874}
\BIBentrySTDinterwordspacing

\bibitem{Other02}
\BIBentryALTinterwordspacing
S.~Shi, X.~Zhang, H.~Li, and W.~Fan, ``Explaining the predictions of any image
  classifier via decision trees,'' \emph{CoRR}, vol. abs/1911.01058, 2019.
  [Online]. Available: \url{http://arxiv.org/abs/1911.01058}
\BIBentrySTDinterwordspacing

\bibitem{Other03}
\BIBentryALTinterwordspacing
W.~Guo, D.~Mu, J.~Xu, P.~Su, G.~Wang, and X.~Xing, ``{LEMNA:} explaining deep
  learning based security applications,'' in \emph{Proceedings of the 2018
  {ACM} {SIGSAC} Conference on Computer and Communications Security, {CCS}
  2018, Toronto, ON, Canada, October 15-19, 2018}, 2018, pp. 364--379.
  [Online]. Available: \url{https://doi.org/10.1145/3243734.3243792}
\BIBentrySTDinterwordspacing

\bibitem{CoreNLP}
\BIBentryALTinterwordspacing
C.~D. Manning, M.~Surdeanu, J.~Bauer, J.~R. Finkel, S.~Bethard, and
  D.~McClosky, ``The stanford corenlp natural language processing toolkit,'' in
  \emph{Proceedings of the 52nd Annual Meeting of the Association for
  Computational Linguistics, {ACL} 2014, June 22-27, 2014, Baltimore, MD, USA,
  System Demonstrations}, 2014, pp. 55--60. [Online]. Available:
  \url{https://doi.org/10.3115/v1/p14-5010}
\BIBentrySTDinterwordspacing

\bibitem{Inception}
C.~{Szegedy}, {Wei Liu}, {Yangqing Jia}, P.~{Sermanet}, S.~{Reed},
  D.~{Anguelov}, D.~{Erhan}, V.~{Vanhoucke}, and A.~{Rabinovich}, ``Going
  deeper with convolutions,'' in \emph{2015 IEEE Conference on Computer Vision
  and Pattern Recognition (CVPR)}, June 2015, pp. 1--9.

\end{thebibliography}

\end{document}